\documentclass[twoside]{article}

\usepackage[accepted]{aistats2022}

\usepackage[round]{natbib}
\renewcommand{\bibname}{References}
\renewcommand{\bibsection}{\subsubsection*{\bibname}}

\usepackage{amsmath,amsfonts,bm}

\def\Figref#1{Figure~\ref{#1}}
\def\twofigref#1#2{figures \ref{#1} and \ref{#2}}

\def\Apxref#1{Appendix~\ref{#1}}

\def\Secref#1{Section~\ref{#1}}

\def\eqref#1{(\ref{#1})}

\def\Algref#1{Algorithm~\ref{#1}}

\def\1{\bm{1}}

\def\rvm{{\boldsymbol{m}}}

\def\rvx{{\boldsymbol{x}}}
\def\rvy{{\boldsymbol{y}}}
\def\rvz{{\boldsymbol{z}}}

\def\mTheta{{\bm\theta}}

\DeclareMathAlphabet{\mathsfit}{\encodingdefault}{\sfdefault}{m}{sl}
\SetMathAlphabet{\mathsfit}{bold}{\encodingdefault}{\sfdefault}{bx}{n}

\def\sM{{\mathbb{M}}}
\def\sN{{\mathbb{N}}}

\newcommand{\E}{\mathbb{E}}

\newcommand{\R}{\mathbb{R}}

\newcommand{\KL}{D_{\mathrm{KL}}}

\DeclareMathOperator*{\argmax}{arg\,max}
\DeclareMathOperator*{\argmin}{arg\,min}

\usepackage{hyperref}
\usepackage{url}
\usepackage{graphicx}
\usepackage{amssymb}
\usepackage{pgfplots}
\usepackage{tikz}
\usepackage{algorithm}
\usepackage{subcaption}
\usepackage{booktabs}
\usepackage{multicol}
\usepackage{float}
\usepackage[export]{adjustbox}
\usetikzlibrary{matrix}
\usetikzlibrary{plotmarks}
\usetikzlibrary{patterns}
\usepackage[noend]{algpseudocode}
\usepackage{helvet}
\usepackage[eulergreek]{sansmath}
\usepackage{setspace}
\usepgfplotslibrary{fillbetween}
\pgfplotsset{
    compat=1.17,
    every tick label/.append style={font=\tiny\sansmath\sffamily},
    every axis label/.append style={font=\tiny\sansmath\sffamily},
    every axis plot/.append style={thick},
    legend style={font=\sansmath\sffamily},
    label style={font=\tiny\sansmath\sffamily},
    legend image code/.code={
        \draw[mark repeat=2,mark phase=2]
        plot coordinates {
            (0cm,0cm)
            (0.25cm,0cm)        
            (0.5cm,0cm)         
        };%
    }
}

\definecolor{cBP}{HTML}{108AE3}
\definecolor{cMF}{HTML}{E3571E}
\definecolor{cdeepBP}{HTML}{1852CC}
\definecolor{cdeepBP8}{HTML}{10AEE3}
\definecolor{cdeepBP32}{HTML}{1CC4DA}
\definecolor{cdeepMF}{HTML}{D62728}
\definecolor{cdeepMF8}{HTML}{ED521F}
\definecolor{cdeepMF32}{HTML}{F69C40}
\definecolor{cMV}{RGB}{44,160,44}
\definecolor{cCL}{HTML}{ED1FD2}
\definecolor{cBayesDGC}{HTML}{A631F5}

\newcommand{\mytimes}{\scalebox{1}{$\times$}}
\newcommand{\mytriangle}{\scalebox{1}{$\triangle$}}
\newcommand{\mybigcirc}{\scalebox{1}{$\bigcirc$}}

\begin{document}

%

%

\twocolumn[

\aistatstitle{Robust Deep Learning from Crowds with Belief Propagation}

\aistatsauthor{ $\text{Hoyoung Kim}^{*}$ \And $\text{Seunghyuk Cho}^{*}$ \And $\text{Dongwoo Kim}$ \And $\text{Jungseul Ok}^{\dagger}$ }

\aistatsaddress{\\ Pohang University of Science and Technology } 
]

\begin{abstract}

Crowdsourcing systems enable us to collect large-scale dataset, but inherently suffer from noisy labels of low-paid workers. We address the inference and learning problems using such a crowdsourced dataset with noise. Due to the nature of sparsity in crowdsourcing, it is critical to exploit both probabilistic model to capture worker prior and neural network to extract task feature despite risks from wrong prior and overfitted feature in practice. We hence establish a neural-powered Bayesian framework, from which we devise deepMF and deepBP with different choice of variational approximation methods, mean field (MF) and belief propagation (BP), respectively. This provides a unified view of existing methods, which are special cases of deepMF with different priors. In addition, our empirical study  suggests that deepBP is a new approach, which is more robust against wrong prior, feature overfitting and extreme workers thanks to the more sophisticated BP than MF.

\end{abstract}

\section{Introduction}

Crowdsourcing systems, such as Amazon Mechanical Turk, 
enable us to collect huge labeled datasets at low budget and in short time
by distributing labeling tasks over the crowd workers.
However, low-paid workers are liable to provide noisy labels,
and even trustworthy workers, called hammers, have non-zero probability to make mistakes.
In addition, there often exist spammers randomly labeling
and adversaries incorrectly labeling due to misinterpretation of task description
or malicious intention~\citep{jagabathula2017identifying}.
Therefore, in order to
fully utilize the crowdsourced dataset from such various workers
for either inferring true labels, i.e., {\it inference from crowds} \citep{dawid1979maximum}, or training a model to perform the same labeling task, i.e., {\it learning from crowds} \citep{raykar2010learning},
it is a fundamental problem to jointly
estimate worker abilities and true labels.
Indeed, with the presence of diverse workers,
majority voting (MV) giving the same weight to each worker often fails at recovering true labels due to spammers or adversaries, whereas a weighted MV giving higher weight to more reliable workers would not if the worker estimation is accurate.


A principled approach for this problem
is to establish a probabilistic generative model on the behaviors of worker labeling
and apply a standard method for inference or learning.
\citet{dawid1979maximum} propose
a pioneering generative model where each worker is associated with confusion matrix,
and an expectation-maximization (EM) algorithm to 
 recursively infer true labels and estimate confusion matrices.
Since then, early works \citep{whitehill2009whose,raykar2010learning, welinder2010multidimensional,liu2012variational}
have developed
sophisticated methods for inference and learning methods 
to be capable of exploiting 
additional information such as a task feature or worker prior.
\citet{raykar2010learning} propose a framework to conduct
inference and learning from crowds in an iterative manner,
where the learning part enables us to utilize task features. 
\citet{liu2012variational} establish a flexible Bayesian approach that allows us to plug-in
any worker prior distribution.

Meanwhile, with the recent advances in deep learning, 
there have been proposed a number of methods
to train neural network model directly from crowdsourced dataset
\citep{rodrigues2018deep, tanno2019learning}.
However, it is well known that
such methods often suffer
from unstable learning mainly since it is inevitable to use local search methods such as a gradient-based optimizer
in deep learning.
The instability issue becomes severe
in canonical scenarios that
the number of workers per task is limited by budget constraints, and each worker has a limited capacity to perform trustworthy labeling, i.e., 
the information in crowdsourced dataset is {\em large but sparse}.\footnote{In \Apxref{appendix:fixed-budget}, we empirically show that given robust algorithms and budget limit,
it is better to obtain large but sparse dataset than small but dense one 
in terms of the performance in learning.}
To mitigate this issue, \citet{tanno2019learning} introduces
a specific regularization term based on the model of worker labeling.
However, 
we empirically found that \citet{tanno2019learning}'s approach 
is sensitive to the initialization or hyperparameter of deep learning\footnote{More details are in \Apxref{appendix:tanno experiment}.}, in particular, when their worker model is mismatched to the real perhaps with outliers.

In this paper, we hence aim at {\em robust} deep learning from crowds.
To do so, 
we first establish a generative model containing a neural network and worker prior
to which we easily insert any worker prior distribution.
This provides not only a flexibility 
but also plausible interpretation to the choice of worker prior.
Using our model,
we then devise 
an EM framework 
alternating variational inference, which is used to infer true label on the probabilistic model, and deep learning, which is used to update the probabilistic model.
In \Secref{sec:deepmf}, 
adopting mean-field (MF) approximation for the varational inference, 
we devise {\em deepMF} as a simple procedure of minimizing a loss function.
We also show that the existing methods can be interpreted as 
an instance of deepMF with a specific choice of worker prior in \Secref{sec:connection}.

Although MF-based approaches are popular thanks to its simplicity \citep{rodrigues2018deep, tanno2019learning},
belief-propagation (BP) is known to be a better approximation method in general \citep{weiss2001comparing,murphy2013machine}.
To be specific, we note that the crowdsourced dataset
can be interpreted as a bipartite weighted graph
consisting of edges between tasks and workers
with the corresponding worker labels as weights.
If the graph is tree, the inference based on BP is exact on the graph~\citep{pearl1982bp}.
Furthermore, \citet{ok2016optimality}
prove the asymptotic optimality of BP
in the canonical scenario where
the assignment graph is a random bipartite graph with limited degrees
so that locally tree-like with high probability.
Intuitively, to infer a task label,
BP is able to fully utilize 
the information spread over the local tree rooted from the task
in a non-backtracking manner~\citep{yedidia2003understanding},
while MF approximation is not precise at this level. 
To take the advantages of BP, in \Secref{sec:deepbp}, we propose \textit{deepBP} 
using BP for the variational inference.


As intended in the design of deepBP, it empirically outperforms deepMF and the other algorithms.
In particular, the advantage of using deepBP is robustness in the set of canonical crowdsourcing scenarios.
We consider the following three specific scenarios, where any workers or practitioners would face in the real world.
First, workers are commonly faced with non-informative or even irrelevant task features in a given task. Consequently, the features become obstacles in learning.
Second, practitioners are likely to choose a wrong worker prior for inference and learning, or the model would not work properly with even true prior.
Third, malicious workers are everywhere, especially the ones who submit a lot of random answers.
All the scenarios are simulated with synthetic data. In all cases, deepBP performs robustly than the other algorithms, although some perform comparably in a particular case but not for all. In an additional experiment on a real-world dataset augmented with spammers and non-informative features, deepBP still maintains its robustness.

\paragraph{Contribution.}
In this work, 
to robustify deep learning from crowds,
we propose a principled framework alternating variational inference and deep learning to utilize both the prior of worker behavior and the features of tasks.
Our contributions are summarized in three folds.
First, we devise deepMF in a simple form of deep learning 
and reveal that the previous methods~\citep{rodrigues2018deep,tanno2019learning} are special cases of deepMF with specific choices of worker prior.
This provides a useful guideline to select algorithms with a plausible interpretation.
Second, inspired by the strong theoretical guarantee on 
BP for inference from crowds \citep{pearl1982bp, ok2016optimality},
we propose deepBP, 
to our best knowledge, which is the first attempt to use BP for deep learning from crowds.
Third, we empirically show that 
deepBP overall outperforms deepMF and the other existing methods.
The advantages of deepBP in terms of robustness is clearer particularly in 
the set of canonical crowdsourcing scenarios 
with i) non-informative features;
ii) multi-modal or mismatched worker prior;
or iii) extreme spammers 
who submit noises to several tasks.

\paragraph{Related Work.}
As mentioned earlier,
sparse dataset is common in crowdsourcing system due to limited budget,
but it is challenging by the risk of overfitting.
As a part of robustifying deep learning on the sparse regime,
our approach is along with the Bayesian view \citep{liu2012variational}
in the sense that the knowledge on workers
is given as a prior distribution of workers' confusion matrices.
Meanwhile, there are diverse ways to express and exploit the worker prior.
Assuming that workers can be clustered into few different types with similar confusion matrix, \citet{venanzi2014community} propose to involve a worker clustering mechanism.
In \citep{nguyen-etal-2017-aggregating},
the confusion matrix is converted into a confusion vector
to stabilize learning by dimensionality reduction.
Interestingly, \citet{chu2021generative} introduce the generative adversarial networks
to learn worker behavior. 
We believe that our approach provides ease of adopting 
different worker prior through the lens of the probabilistic perspective.


As deepBP is a deep learning method equipped with BP,
it is worth to mention studies to take advantages of BP and deep learning
for other learning problems.
To capture high order dependencies on factor graphs, \citet{NEURIPS2020_61c66a2f} propose a factor graph neural network which can mimic the max-product BP.
\citet{NEURIPS2020_07217414} generalize BP with neural network to find better fixed posteriors faster than loopy BP.
Recently, \citet{satorras2021neural} propose a hybrid model which runs conjointly a graph neural network with BP.
The above methods are designed
for flexible or fast inference 
by training neural network from multiple instances of inference on a {\em fixed} factor graph.
However, in crowdsourcing systems, the factor graph varies easily per each instance.
We hence need to devise a new framework alternating inference and learning.

\section{Model}



We consider a classification model to predict 
latent label $z \in [K]: = \{1, 2, ..., K\}$
from feature $x$ such as an image or audio track.
We model the probability of class $k$ given feature $x$ 
using an arbitrary function $f_{\phi}(k; x)$ parameterized by $\phi$,
i.e.,
\begin{equation}
\label{eqn:task_model}
p(z = k \mid x, \phi) = f_\phi(k; x) \;,
\end{equation}
which can be logistic regression model as in \citep{raykar2010learning},
or deep neural network as in \citep{rodrigues2018deep}.

To collect training dataset, 
we consider a crowdsourcing model
consisting of $N$ classification tasks and $M$ workers.
Each task $i \in [N]$ 
is associated with true label $z_i \in [K]$
and feature $x_i$. We assume each task is sampled independently,
i.e., 
$p(\bm{z} \mid \bm{x}, \phi)= \prod_{i \in [N]} p(z_i \mid x_i, \phi)$, 
where $\bm{z}$ and $\bm{x}$ are the sets of all $z_i$'s and $x_i$'s, respectively.
%
Each worker $u \in [M]$ is associated with 
confusion matrix $\bm\theta^{(u)} \in [0,1]^{K \times K}$, parameterizing worker $u$'s average ability, such that
$\theta^{(u)}_{kk'}$ is 
the probability of answering $k'$ for tasks with true label $k$:
\begin{equation}
\label{eqn:confusion}
p(y^{(u)}_{i} = k' \mid z_i = k, \bm\theta^{(u)}) = \theta^{(u)}_{kk'}\;,
\end{equation}
where $y^{(u)}_{i} \in [K]$ is worker $u$'s answer for task $i$,
and $\sum_{k' \in [K]} \theta^{(u)}_{kk'}  = 1$.

We assume that
each confusion matrix $\bm\theta^{(u)}$
is drawn independently from 
prior distribution $p(\bm\theta^{(u)} \mid \bm\alpha)$
with parameter $\bm{\alpha}$, i.e.,
$p(\bm{\theta} \mid  \bm\alpha) =
\prod_{u \in [M]} p(\bm{\theta}^{(u)} \mid  \bm\alpha)$,
where $\bm{\theta}$ is the set of all $\bm{\theta}^{(u)}$'s.
The worker prior $p(\bm{\theta}^{(u)} \mid  \bm\alpha)$
is often equipped with 
Dirichlet distribution $\text{Dir}(\bm\alpha)$
since it provides an analytical tractability 
from the fact that it is a conjugate prior of the categorical distribution, 
and also represents a wide set of distributions by simply manipulating $\bm\alpha$.
For instance, in \citep{liu2012variational},
the one-coin model for binary classification 
uses the worker prior such that for $\alpha_1, \alpha_2 > 0$,
\begin{equation} \label{eq:one-coin-prior}
(\theta^{(u)}_{11}, \theta^{(u)}_{12})=
(\theta^{(u)}_{22}, \theta^{(u)}_{21}) 
\;{\sim}_{\text{i.i.d.}}\; \text{Dir} (\alpha_{1}, \alpha_{2})\;,
\end{equation}
where the expected probability of being correct is
$\frac{\alpha_1}{\alpha_1  + \alpha_2}$.
The two coin model in \citep{liu2012variational}
independently draws
$(\theta^{(u)}_{11}, \theta^{(u)}_{12})$
and  $(\theta^{(u)}_{22}, \theta^{(u)}_{21})$
from the same distribution.

We assume that 
$y^{(u)}_i$'s are conditionally independent to each other
given $\bm{z}$ and $\bm{\theta}$. 
Let $\bm{y}$ be the set of all $y^{(u)}_i$'s, and 
$\sN_u$ be the set of tasks labeled by worker $u$.
The model described above can be summarized as:
\begin{align}
& p(\rvy, \rvz, \bm\theta \mid \rvx, \bm\alpha, \phi)= p(\rvz \mid \rvx, \phi) p(\rvy \mid \rvz, \mTheta) p(\mTheta \mid \bm\alpha) \nonumber \\
&\!= 
\!\! \prod_{i\in [N]} \! f_\phi (z_i;x_i) 
\left( \prod_{u\in[M]} p(\bm\theta^{(u)} | \bm{\alpha}) \prod_{j \in \sN_u} \theta^{(u)}_{z_j, y^{(u)}_j} \right)
\;.
\label{eqn:joint_learning}
\end{align}
Given this model
equipped with
task feature $f_\phi(z_i ; x_i)$
and worker prior $p(\bm\theta^{(u)} \mid \bm{\alpha})$,
in what follows, we describe 
two fundamental problems: inference and learning from crowds.

\paragraph{Inference from crowds.}
Given crowdsourced dataset $(\rvx, \rvy)$,
worker prior $\bm\alpha$, and classifier $f_\phi$ based on task feature, 
the Bayes decision rule
is given as:
\begin{equation}
\label{eq:inference}
\hat z_i = \argmax_{z_i \in [K]} p(z_i \mid \rvx, \rvy, \bm\alpha, \phi) \;,
\end{equation}
which is optimal in the sense of minimizing the expected bit-wise error rate.
Noting $p(\rvz, \mTheta \mid \rvx, \rvy, \bm\alpha, \phi) 
\propto p(\rvz \mid \rvx, \phi) p(\rvy \mid \rvz, \mTheta) p(\mTheta \mid \bm\alpha)$,
the marginal probability 
in \eqref{eq:inference}
can be computed as follows:
\begin{align}
& p(z_i \mid \rvx, \rvy, \bm\alpha, \phi)
= \sum_{\rvz_{\text{-}i}} \int p(\rvz, \mTheta \mid \rvx, \rvy, \bm\alpha, \phi) d\mTheta \nonumber
\\
&\propto
\sum_{\rvz_{\text{-}i}}
 p(\rvz \mid \rvx, \phi) 
 \int p(\rvy \mid \rvz, \mTheta) p(\mTheta \mid \bm\alpha) d\mTheta \;,
\label{eq:marginalization-infer}
\end{align}
where $\rvz_{\text{-}i} := (z_j:  i \neq j \in [N] )$.
In general, the marginalization is computationally intractable
since the summation in \eqref{eq:marginalization-infer}
takes over exponentially many 
$\rvz_{\text{-}i} \in [K]^{N-1}$. 
To approximate
such an intractable marginalization, 
one can apply a variational method such as MF or BP.

\begin{figure*}[!t]
\begin{minipage}{0.49\textwidth}
\begin{algorithm}[H]
\setstretch{1.35}
\caption{deepMF($\rvx, \rvy, \phi, \bm\alpha, \mTheta, \bm\beta, c$)}
\begin{algorithmic}[1]
\While {not converged}
    \State $\hat{f}_{\phi}(\rvx) \gets \text{Clip}\left(f_{\phi}(\rvx), c \right)$
    \State $q(\rvz) \gets \text{MF} \left( \hat{f}_{\phi}(\rvx), \rvy, \bm{\alpha}, \mTheta, \bm\beta \right)$
    \State $\phi, \bm\beta \gets \argmax_{\phi,\bm\beta} \mathcal{L}_{\text{MF}}\left(q(\rvz) ; \phi, \mTheta, \bm\beta \right)$
\EndWhile
    \State \Return $q(\rvz), \phi$
\end{algorithmic}
\label{alg:MF}
\end{algorithm}
\end{minipage}
\hfill
\begin{minipage}{0.49\textwidth}
\begin{algorithm}[H]
\setstretch{1.35}
\caption{deepBP($\rvx, \rvy, \phi, \bm\alpha, c$)}
\begin{algorithmic}[1]
\While {not converged}
    \State $\hat{f}_{\phi}(\rvx) \gets \text{Clip}\left(f_{\phi}(\rvx), c \right)$
    \Comment{Clipping}
    \State $q(\rvz) \gets \text{BP}\left( \hat{f}_{\phi}(\rvx) , \rvy, \bm\alpha \right)$ \Comment{Inference}
    \State $\phi \gets \argmax_\phi \mathcal{L}_{\text{BP}}\left(q(\rvz) ; \phi \right)$ \Comment{Learning}
\EndWhile
    \State \Return $q(\rvz), \phi$
\end{algorithmic}
\label{alg:BP}
\end{algorithm}
\end{minipage}
\end{figure*}

\paragraph{Learning from crowds.}
Given crowdsourced dataset $(\rvx, \rvy)$
and worker prior $\bm\alpha$, 
we can formulate 
the problem of learning classifier $f_\phi$
as the maximization of the posterior: 
\begin{equation}
\hat\phi = \argmax_\phi p(\phi \mid \rvx, \rvy, \bm\alpha) \;.
\label{eq:learning}
\end{equation}
The maximum a posterior (MAP) estimate $\hat{\phi}$
can be used for not only
the optimal inference in \eqref{eq:inference},
but also for predicting label $z_{*}$ of unseen $x_{*}$ as follows:
\begin{equation}
\hat{z}_{*} = \argmax_{k \in [K]} f_{\phi} (z_{*} = k; x_{*}) \;.
\end{equation}

Using the Bayes rule, the posterior is written as:
\begin{equation}
p(\phi \mid \rvx, \rvy, \bm\alpha)
\propto p(\rvy \mid \rvx, \bm\alpha, \phi) p(\phi) \;,
\end{equation}
where if the prior $p(\phi)$ is assumed to be zero-mean Gaussian, then it will be translated into the negative of
L2-norm regularizer of $\phi$ in \eqref{eq:learning}.
Hence, to obtain the MAP estimate 
in \eqref{eq:learning},
we write the likelihood $p(\rvy \mid \rvx, \bm\alpha, \phi)$
as follows:
\begin{align}
p(\rvy \mid \rvx, \bm\alpha, \phi)
&= \sum_{\rvz} p(\rvy, \rvz \mid \rvx, \bm\alpha, \phi) \label{eq:one margin} \\
&= \sum_{\rvz} \int p(\rvy, \rvz, \mTheta \mid \rvx, \bm\alpha, \phi) d\mTheta \;, \label{eq:two margin}
\end{align}
where analog to \eqref{eq:marginalization-infer},
in general,
it is intractable to marginalizing out $\rvz$ 
due to the exponentially many summations 
over $\rvz \in [K]^N$.

\section{Method} \label{sec:method}
We propose deepMF and deepBP, 
each of which is essentially an expectation-maximization (EM) algorithm consisting of iterations of E- and M-steps addressing
the intractable marginalization issues for
the inference \eqref{eq:inference} and the learning \eqref{eq:learning}, respectively,
with different variational Bayesian approaches.
To be specific, 
E-step estimates $q_i(z_i)$
to approximate  $p(z_i \mid \rvx, \rvy, \bm\alpha, \phi)$
in \eqref{eq:marginalization-infer},
where deepMF and deepBP use MF approximation and BP algorithm, respectively.
In M-step, given $q(\rvz)$ from E-step,
classifier $f_\phi$ is trained to
maximize an evidence lower bound (ELBO)
of the likelihood $p(\rvy \mid \rvx, \bm\alpha, \phi)$,
derived from \eqref{eq:one margin} 
and \eqref{eq:two margin} for deepMF and deepBP, correspondingly.
In what follows,
we first present formal descriptions of 
deepMF (Section~\ref{sec:deepmf}) and deepBP (Section~\ref{sec:deepbp}),
and then in Section~\ref{sec:method-remark}, we provide useful remarks.





\subsection{Deep Mean-Field}
\label{sec:deepmf}
We derive deepMF from an ELBO of the marginalization in \eqref{eq:two margin}
by introducing a variational distribution $q(\rvz,\mTheta)$ in the followings:
\begin{align}
&\log p(\rvy \mid \rvx, \bm{\alpha}, \phi) \nonumber
= \log \E_{q(\rvz, \mTheta)} \left[ \frac{p(\rvy, \rvz, \mTheta \mid \rvx, \bm{\alpha}, \phi)}{q(\rvz, \mTheta)} \right] \nonumber \\
& \geq \E_{q(\rvz, \mTheta)} \left[ \log \frac{p(\rvy, \rvz, \mTheta \mid \rvx, \bm{\alpha}, \phi)}{q(\rvz, \mTheta)} \right] \;.
\label{eq:joint elbo}
\end{align}
Note that the lower bound is maximized and achieves the equality 
if $q(\rvz,\mTheta) = p(\rvz, \mTheta \mid \rvx, \rvy, \bm\alpha, \phi)$.
We apply MF approximation,
assuming a conditional independence 
of $z_i$'s and $\mTheta^{(u)}$'s given $(\rvx, \rvy, \bm\alpha, \phi)$,
such that:
\begin{equation}
\!\! q(\rvz, \mTheta) \! = \! q(\rvz) q(\mTheta; \bm\beta) \! = 
\!\!\! \prod_{i\in[N]} \!\!\! q_i(z_i) \!\!\!\!\prod_{u\in[M]} 
\!\!\! q_u(\mTheta^{(u)}; \bm\beta^{(u)}) \!\!\;,
\label{eq:mf_variational}
\end{equation}
where
we use a parametric estimation $q_u(\mTheta^{(u)}; \bm\beta^{(u)}) = \text{Dir} (\mTheta^{(u)}; \bm\beta^{(u)})$ for each worker $u$,
and 
approximating
$q_i(z_i) \approx p(z_i | \rvx, \rvy, \bm\alpha, \phi)$
and $q_u(\mTheta^{(u)}; \bm\beta^{(u)}) \approx p(\mTheta^{(u)} | \rvx, \rvy, \bm\alpha, \phi)$,
we let
$q(\rvz) = \prod_{i\in[N]} q_i(z_i)$ and $q(\mTheta)  =\prod_{u\in[M]} q_u(\mTheta^{(u)})$.
Then, using \eqref{eqn:joint_learning} and \eqref{eq:mf_variational}, the ELBO in \eqref{eq:joint elbo} is written as follows:
\begin{align}
& \mathcal{L}_{\text{MF}}(q(\rvz) ; \phi, \mTheta, \bm\beta)
:= \E_{q(\rvz) q(\mTheta; \bm\beta)} [\log p(\rvy \mid \rvz, \mTheta)] \nonumber \\
& \! - \! \KL(q(\rvz) || p(\rvz | \rvx, \phi)) \!
- \! \KL(q(\mTheta; \bm\beta) || p(\mTheta | \bm{\alpha})) \;.
\label{eq:deepMF}
\end{align}

\begin{table}[t!]
    \setstretch{1.15}
    \centering
    \caption{{\em Comparison between deepMF and the previous methods:
    CL \citep{rodrigues2018deep}, Trace \citep{tanno2019learning}, BayesDGC \citep{li2021crowdsourcing}.
    } \texttt{Prior} indicates the use of worker specific prior in each model. \texttt{Variational} indicates the use of variational methods in the parameter estimation. \texttt{Analytic $\beta$} shows the presence of an analytic solution for the point-wise estimation of $\beta$.
    }
    \resizebox{\linewidth}{!}{
    \begin{tabular}{l c c c}
        \toprule
        & \!\!{\tt Prior}\!
        & \!{\tt Variational}\!
         & 
        \!{\tt Analytic $\beta$}\!
        \\
        \midrule
        CL & $\mytimes$ & $\mytimes$ & $\mytimes$ \\
        Trace & $\mytriangle$ & $\mytimes$ & $\mytimes$ \\
        BayesDGC & $\mybigcirc$ & $\mybigcirc$ & $\mytimes$ \\
        deepMF (ours) & $\mybigcirc$ & $\mybigcirc$ & $\mybigcirc$ \\
        \bottomrule
    \end{tabular}
    }
    \label{table:existing methods}
\end{table}

In each iteration of deepMF,
we seek $q(\rvz)$, $\bm\beta$ and $\phi$, sequentially, 
to maximize $\mathcal{L}_{\text{MF}}(q(\rvz) ; \phi, \mTheta, \bm\beta)$
by fixing the others. 
Recalling $p(\rvz \mid \rvx, \phi) := \prod_{i \in [N]} f_\phi (z_i ; x_i)$,
given
the classifier $f_\phi$ and
the other variational distributions, 
the ELBO is maximized at $q_i (z_i)$ such that:
\begin{align}
&\log q_i (z_i)  \label{eq:inference mf} \\
&=  \sum_{u \in \sM_i} \E_{q_u(\mTheta^{(u)}; \bm\beta^{(u)})} \big[\log
\theta^{(u)}_{z_i, y_{i}^{(u)}}\big] 
+ \log f_\phi (z_i ; x_i) - 1 \;, \nonumber
\end{align}
where $\sM_i$ denotes the workers who label task $i$.
The collection is denoted by $\text{MF} ({f}_{\phi}(\rvx), \rvy, \bm{\alpha}, \mTheta, \bm\beta )$.
The inference of deepMF
is based on $q_i(z_i)$ from $\text{MF}(\cdot)$.

Then, provided $q(\rvz)$ from $\text{MF}(\cdot)$, the ELBO is maximized at $q_u(\mTheta^{(u)};\bm\beta)$ such that:
\begin{align}
&\log q_u(\mTheta^{(u)};\bm\beta) \\
&= \sum_{i \in \sN_u} \E_{q(\rvz_i)} \left[\log p(y_{i}^{(u)} | \rvz_i, \mTheta^{(u)})\right]
+ \log p(\mTheta^{(u)} | \bm\alpha) - 1 \;. \nonumber
\label{eq:concentration}
\end{align}
When Dirichlet distribution is used for worker prior, i.e., $p(\mTheta^{(u)} | \bm\alpha) := \text{Dir}(\mTheta^{(u)};\bm\alpha)$, the parameter $\bm\beta$ is updated as:
\begin{equation}
\beta^{(u)}_{k_1k_2} = \alpha_{k_1k_2} + \sum_{\{i \mid i \in \sN_u, y^{(u)}_i = \, k_2\}}q_i(z_i=k_1) \;, \label{eq:beta}
\end{equation}
for all $k_1, k_2 \in [K]$. 
Finally, deepMF finds $\phi$ independently of $\bm\beta$ by maximizing the ELBO in \eqref{eq:deepMF}:
\begin{equation}
\hat{\phi} = \argmax_{\phi} \mathcal{L}_{\text{MF}}(q(\rvz) ; \phi, \mTheta, \bm\beta) \;,
\label{eq:learning mf}
\end{equation}
which corresponds to training classifier $f_\phi$.

We note that deepMF is easily implementable as an iteration of the inference \eqref{eq:inference mf} and learning \eqref{eq:learning mf}
is just a sequence of maximizations for the same objective \eqref{eq:deepMF} but different variables.
Hence, it is a popular framework despite the coarse approximation in \eqref{eq:mf_variational}. 
The overall procedure of deepMF is summarized in \Algref{alg:MF},
and the detailed equations of deepMF can be found in \Apxref{appendix:deepMF}.

\subsection{Connection to existing methods.}
\label{sec:connection}
\citet{rodrigues2018deep} and \citet{tanno2019learning} are specific instances of deepMF with different choices of variational distributions and worker prior $\bm\alpha$ in \eqref{eq:deepMF}.
Both \citet{rodrigues2018deep} and \citet{tanno2019learning} use a neural network parameterized by $\phi$ as a variational distribution to approximate the posterior of the true label and the Dirac delta function to confusion matrix $q(\mTheta)$, i.e., point-wise estimation on the confusion matrix.
With the variational distributions, the ELBO in \eqref{eq:deepMF} is simplified as:
\begin{equation}
\label{eq:cl}
\!\!\!\mathcal{L}(\rvz; \phi, \mTheta)
\!:=\! \E_{p(\rvz \mid \rvx, \phi)} [\log p(\rvy \!\mid\! \rvz, \mTheta)] + \log p(\mTheta \!\mid\! \bm\alpha) \;.
\end{equation}
\citet{rodrigues2018deep} further assumes that the ability of workers follows the uniform distribution by fixing all $\alpha_i$ to $1$.
Separately, \citet{tanno2019learning} assumes the presence of more adversaries than hammers using an adversarial prior where $\alpha_{k_1 k_2}$ less than one when $k_1 = k_2$ and one otherwise.
In this work, we do not predetermine the hyperparameter, and instead, we show the robustness of the proposed framework to the choice of hyperparameters.
Furthermore, deepMF approximates the distribution over the confusion matrix by placing a Dirichlet variational distribution. 

BayesDGC~\citep{li2021crowdsourcing}
combines deep neural networks and probabilistic graphical model and is the most similar to deepMF.
BayesDGC uses a natural-gradient stochastic variational inference combining variational message passing and stochastic gradient descent.
Unlike BayesDGC, we derive a closed form update for parameter $\bm\beta$.
The comparison between our methods and previous approaches is summarized in Table \ref{table:existing methods} and \Apxref{appendix:MF-based algorithms} 

\subsection{Deep Belief-Propagation}
\label{sec:deepbp}
To derive deepBP,
we start with an ELBO for the alternative marginalization in \eqref{eq:one margin}
using a variational distribution $q(\rvz)$ as follows:
\begin{align}
& \log p(\rvy \mid \rvx, \bm\alpha, \phi) \nonumber
= \log \E_{q(\rvz)} \left[ \frac{p(\rvy, \rvz \mid \rvx, \bm\alpha, \phi)}{q(\rvz)} \right] \nonumber \\
& \geq \E_{q(\rvz)} \left[ \log \frac{p(\rvy, \rvz \mid \rvx, \bm\alpha, \phi)}{q(\rvz)} \right]\;, 
\label{eqn:first_bp_elbo}
\end{align}
where the lower bound is maximized with the equality 
when $q(\rvz) = p(\rvz \mid \rvx, \rvy, \bm\alpha, \phi)$.
In deepBP, we again recursively update $q(\rvz)$ and $\phi$,
but employ BP to compute $q(\rvz)$.
Unlike the MF approximation, it is known that BP computes the exact posterior when tree structure of the assignment graph is given~\citep{pearl1982bp}.
Furthermore, without the features $\rvx$, the inference with BP is exactly optimal under some mild assumptions~\citep{ok2016optimality}.

We first describe the use of BP for $q(\rvz)$. To do so,
using \eqref{eqn:joint_learning},
we correspond the complete-data likelihood 
in \eqref{eqn:first_bp_elbo} to the following factor graph form:
\begin{align}
&p(\rvy, \rvz \mid \rvx, \bm{\alpha}, \phi) 
= \int p(\rvy, \rvz, \mTheta \mid \rvx, \bm{\alpha}, \phi) \, d\mTheta  \label{eq:factor_graph} \\
& \propto \prod_{i\in[N]}
\underbrace{ 
\vphantom{ \int_{\mTheta^{(u)}} p(\mTheta^{(u)} \mid \bm{\alpha}) \prod_{j \in \sN_u} \theta^{(u)}_{z_j,y^{(u)}_j} d\mTheta^{(u)}  } 
f_\phi(z_i ; x_i) }_{=: h_i(z_i;\phi)} \prod_{u\in[M]} 
\underbrace{ \int p(\mTheta^{(u)} | \bm{\alpha}) \prod_{j \in \sN_u} \theta^{(u)}_{z_j,y^{(u)}_j} d\mTheta^{(u)}  }_{=: g_u(\rvz_{\sN_u};\bm\alpha)} \;, \nonumber 
\end{align}
where 
we let 
$h_i(z_i;\phi)$ be the task $i$'s factor
containing feature information 
and $g_u(\rvz_{\sN_u};\bm\alpha)$ 
be the worker $u$'s factor including worker prior.
Given $f_\phi$,
the sum-product BP on the factor graph
can be written as the following iterative updates 
of three types of messages:
\begin{align}
m^{t+1}_{i \rightarrow g_u}(z_i) &\propto m_{h_i \rightarrow i}(z_i) \prod_{v \in \sM_i \backslash \{u\}} m^{t}_{g_v \rightarrow i}(z_i) \;, \nonumber \\
m^{t+1}_{g_u \rightarrow i}(z_i) &\propto \sum_{z_{\sN_u \backslash \{i\}}} g_u(\rvz_{\sN_u}) \prod_{j \in \sN_u} m^{t+1}_{j \rightarrow g_u}(z_j) \;, \label{eq:worker_to_task} \\
m_{h_i \rightarrow i}(z_i) &= f_{\phi}(z_i ; x_i) \;, \nonumber
\end{align}
where the superscription $t$ denotes the index of BP iteration and is omitted for $m_{h_i \rightarrow i}$ as it has no change unless $f_\phi$ is updated. 
We take the standard initialization of the worker-to-task message at uniform distribution, i.e., $m^0_{g_u \rightarrow i} = [1/K]^{K}$.
After updating the messages $T$ iterations,
we use
\begin{equation}
q_i(z_i) \propto m_{h_i \rightarrow i}(z_i) \prod_{u \in \sM_i} m^T_{g_u \rightarrow i}(z_i) \;,
\end{equation}
where $\text{BP} ({f}_{\phi}(\rvx), \rvy, \bm{\alpha})$ denotes the entire collections
of $q_i(z_i)$ from BP.
The inference of deepBP is directly based on $q_i(z_i)$ from $\text{BP}(\cdot)$.

We now describe the update of classifier $f_\phi$.
Using \eqref{eq:factor_graph}, the ELBO in \eqref{eqn:first_bp_elbo} can be simplified as:
\begin{align}
&\mathcal{L}_{\text{BP}}(q(\rvz);\phi)
:= \E_{q(\rvz)} \left[ \log \frac{h(\rvz;\phi) g(\rvz;\bm\alpha)}{q(\rvz)} \right] \nonumber \\
&= \E_{q(\rvz)} \left[ \log g(\rvz;\bm\alpha) \right] - \KL(q(\rvz) \parallel h(\rvz;\phi)) \;,
\label{eq:bp_elbo}
\end{align}
where 
$h(\rvz;\phi) := \prod_{i\in[N]} h_i(z_i;\phi)$ 
and $g(\rvz;\bm\alpha) := \prod_{u\in[M]} g_u(\rvz_{\sN_u};\bm\alpha)$.
Then, given $q(\rvz)$ from $\text{BP}(\cdot)$, 
deepBP obtains $\phi$ to maximize the ELBO in \eqref{eq:bp_elbo}:
\begin{equation}
\hat{\phi} = \argmax_{\phi} \mathcal{L}_{\text{BP}}(q(\rvz) ; \phi) \;.
\label{eq:learning bp}
\end{equation}
The overall process of deepBP is described at \Algref{alg:BP}.


\subsection{Implementation Remarks}
\label{sec:method-remark}

\paragraph{Clipping trick.}
In both deepMF and deepBP, 
the deep learning step can be unstable in early phase in particular when we train neural network from scratch. In addition, we often find overfitting in deep learning.
These observation motivates us to regulate the influence
of deep classifier $f_\phi$ in the inference step.
To do so, to update $q(\rvz)$ in both algorithms, 
we clip the output of $f_\phi(\cdot)$ at $c$
and then evenly distribute 
the clipped amount to the other classes
for normalization, i.e., for each class $k$,
we have the clipped output:
\begin{equation}
\text{Clip}(f_{\phi}(k;\rvx), c) \le c ~~~ \forall k \in [K] \;.
\end{equation}
Then,  the clipping parameter $c$ can be adjusted accordingly to the certainty on features. 
For instance of classification from severely blurred images, 
we may set $c$ close to $1/K$ to cope overfitting.




\paragraph{Fast BP message update.}
We remark that the update of $m^{t+1}_{g_u \rightarrow i}$ in \eqref{eq:worker_to_task}
seems intractable as it requires exponentially many summations in the number of tasks labeled by worker, i.e., $O(2^{|\sN_u|})$.
To bypass the intractable computation,
we use a Monte-Carlo method using $S$ samples,
of which computational cost is bounded by $O(|\sN_u| \cdot K \cdot S)$.
The detailed derivation and the performance with respect to the sample size can be found in \Apxref{appendix:bp_optimization}. We note that for a wide family of worker prior,
\citet{liu2012variational} propose another method with divide $\&$ conquer and fast Fourier transformation of $O(|\sN_u| \log^2 |\sN_u| \cdot K^2)$ complexity.
\citet{liu2012variational}'s method is exact for the prior family. 
However, our method is universal to any prior distribution, but also 
is simply implementable with the utilization of GPU's parallel computing.

\section{Numerical Analysis}

We evaluate our algorithms, deepMF and deepBP, on inference and learning tasks for a image classification on both synthetic and real-world datasets. We compare the proposed algorithms with learning algorithms: CL~\citep{rodrigues2014gaussian}, BayesDGC~\citep{li2021crowdsourcing} and classic inference algorithms: MV, MF and BP~\citep{liu2012variational}.
In the inference task, we report the accuracy of the estimated true labels obtained from \eqref{eq:inference}.
In the learning task, we report the accuracy of the classifier on unseen data.
For classifier $f_\phi$, we use a four-layer convolutional network and train with an Adam optimizer.
An average outcome with 50 different random seeds and the 99\% confidence interval are reported for all experiments.
More details on settings can be found in \Apxref{appendix:exp-details}.

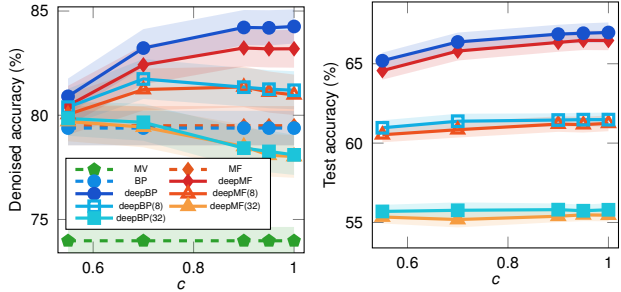
\begin{figure}[t!]
    \captionsetup[subfigure]{font=scriptsize,labelfont=scriptsize,aboveskip=0.05cm,belowskip=-0.15cm}
    \centering
    \begin{subfigure}[t!]{.49\linewidth}
        \centering
        \begin{tikzpicture}
            \begin{axis}[
                legend style={nodes={scale=0.35}, at={(0.03, 0.24)}, anchor=west}, 
                xlabel={$c$},
                ylabel={Denoised accuracy (\%)},
                width=1.2\linewidth,
                height=1.2\linewidth,
                ymin=73.4,
                ymax=85.2,
                xlabel style={yshift=0.15cm},
                ylabel style={yshift=-0.15cm},
                legend columns=2,
                xmin=0.53,
                xmax=1.02,
            ]
                \addplot[cMV, very thick, mark=pentagon*, dashed, mark size=2pt, mark options={solid}] table[col sep=comma, x=x, y=mv]{data/blur_inference.csv};
                \addplot[cMF, very thick, mark=diamond*, dashed, mark size=2pt, mark options={solid}] table[col sep=comma, x=x, y=mf]{data/blur_inference.csv};
                \addplot[cBP, very thick, mark=*, dashed, mark size=2pt, mark options={solid}] table[col sep=comma, x=x, y=bp]{data/blur_inference.csv};
                \addplot[cdeepMF, very thick, mark=diamond*, mark size=2pt, mark options={solid}] table[col sep=comma, x=x, y=deepmf2]{data/blur_inference.csv};
                \addplot[cdeepBP, very thick, mark=*, mark size=2pt, mark options={solid}] table[col sep=comma, x=x, y=deepbp2]{data/blur_inference.csv};
                \addplot[cdeepMF8, very thick, mark=triangle, mark size=2.5pt, mark options={solid}] table[col sep=comma, x=x, y=deepmf8]{data/blur_inference.csv};
                \addplot[cdeepBP8, very thick, mark=square, mark size=2pt, mark options={solid}] table[col sep=comma, x=x, y=deepbp8]{data/blur_inference.csv};
                \addplot[cdeepMF32, very thick, mark=triangle*, mark size=2.5pt, mark options={solid}] table[col sep=comma, x=x, y=deepmf32]{data/blur_inference.csv};
                \addplot[cdeepBP32, very thick, mark=square*, mark size=2pt, mark options={solid}] table[col sep=comma, x=x, y=deepbp32]{data/blur_inference.csv};

                \addplot[name path=deepbpd2,draw=none,fill=none] table[col sep=comma, x=x, y=deepbpd2]{data/blur_inference.csv};
                \addplot[name path=deepbpu2,draw=none,fill=none] table[col sep=comma, x=x, y=deepbpu2]{data/blur_inference.csv};
                \addplot[cdeepBP,fill opacity=0.15] fill between[of=deepbpd2 and deepbpu2];
                
                \addplot[name path=deepbpd8,draw=none,fill=none] table[col sep=comma, x=x, y=deepbpd8]{data/blur_inference.csv};
                \addplot[name path=deepbpu8,draw=none,fill=none] table[col sep=comma, x=x, y=deepbpu8]{data/blur_inference.csv};
                \addplot[cdeepBP8,fill opacity=0.15] fill between[of=deepbpd8 and deepbpu8];
                
                \addplot[name path=deepbpd32,draw=none,fill=none] table[col sep=comma, x=x, y=deepbpd32]{data/blur_inference.csv};
                \addplot[name path=deepbpu32,draw=none,fill=none] table[col sep=comma, x=x, y=deepbpu32]{data/blur_inference.csv};
                \addplot[cdeepBP32,fill opacity=0.15] fill between[of=deepbpd32 and deepbpu32];
                
                \addplot[name path=deepmfd2,draw=none,fill=none] table[col sep=comma, x=x, y=deepmfd2]{data/blur_inference.csv};
                \addplot[name path=deepmfu2,draw=none,fill=none] table[col sep=comma, x=x, y=deepmfu2]{data/blur_inference.csv};
                \addplot[cdeepMF,fill opacity=0.15] fill between[of=deepmfd2 and deepmfu2];
                
                \addplot[name path=deepmfd8,draw=none,fill=none] table[col sep=comma, x=x, y=deepmfd8]{data/blur_inference.csv};
                \addplot[name path=deepmfu8,draw=none,fill=none] table[col sep=comma, x=x, y=deepmfu8]{data/blur_inference.csv};
                \addplot[cdeepMF8,fill opacity=0.15] fill between[of=deepmfd8 and deepmfu8];
                
                \addplot[name path=deepmfd32,draw=none,fill=none] table[col sep=comma, x=x, y=deepmfd32]{data/blur_inference.csv};
                \addplot[name path=deepmfu32,draw=none,fill=none] table[col sep=comma, x=x, y=deepmfu32]{data/blur_inference.csv};
                \addplot[cdeepMF32,fill opacity=0.15] fill between[of=deepmfd32 and deepmfu32];
                
                \addplot[name path=mvd,draw=none,fill=none] table[col sep=comma, x=x, y=mvd]{data/blur_inference.csv};
                \addplot[name path=mvu,draw=none,fill=none] table[col sep=comma, x=x, y=mvu]{data/blur_inference.csv};
                \addplot[cMV, fill opacity=0.15] fill between[of=mvd and mvu];
                
                \addplot[name path=mfd,draw=none,fill=none] table[col sep=comma, x=x, y=mfd]{data/blur_inference.csv};
                \addplot[name path=mfu,draw=none,fill=none] table[col sep=comma, x=x, y=mfu]{data/blur_inference.csv};
                \addplot[cMF, fill opacity=0.15] fill between[of=mfd and mfu];
                
                \addplot[name path=bpd,draw=none,fill=none] table[col sep=comma, x=x, y=bpd]{data/blur_inference.csv};
                \addplot[name path=bpu,draw=none,fill=none] table[col sep=comma, x=x, y=bpu]{data/blur_inference.csv};
                \addplot[cBP, fill opacity=0.15] fill between[of=bpd and bpu];
            
                \legend{MV,MF,BP,deepMF,deepBP,deepMF(8),deepBP(8),deepMF(32),deepBP(32)}
            \end{axis}
        \end{tikzpicture}
        \caption{Inference with blurred images}
        \label{fig:blur_inference}
    \end{subfigure}
    \hspace{-1mm}
    \begin{subfigure}[t!]{.49\linewidth}
        \centering
        \begin{tikzpicture}
            \begin{axis}[
                legend style={nodes={scale=0.5, transform shape}},
                xlabel={$c$},
                ylabel={Test accuracy (\%)},
                width=1.2\linewidth,
                height=1.2\linewidth,
                xlabel style={yshift=0.15cm},
                ylabel style={yshift=-0.15cm},
                legend pos=outer north east,
                xmin=0.53,
                xmax=1.02
            ]
                \addplot[cdeepMF, very thick, mark=diamond*, mark size=2pt, mark options={solid}] table[col sep=comma, x=x, y=deepmf2]{data/blur_learning.csv};
                \addplot[cdeepMF8, very thick, mark=triangle, mark size=2.5pt, mark options={solid}] table[col sep=comma, x=x, y=deepmf8]{data/blur_learning.csv};
                \addplot[cdeepMF32, very thick, mark=triangle*, mark size=2.5pt, mark options={solid}] table[col sep=comma, x=x, y=deepmf32]{data/blur_learning.csv};
                \addplot[cdeepBP, very thick, mark=*, mark size=2pt, mark options={solid}] table[col sep=comma, x=x, y=deepbp2]{data/blur_learning.csv};
                \addplot[cdeepBP8, very thick, mark=square, mark size=2pt, mark options={solid}] table[col sep=comma, x=x, y=deepbp8]{data/blur_learning.csv};
                \addplot[cdeepBP32, very thick, mark=square*, mark size=2pt, mark options={solid}] table[col sep=comma, x=x, y=deepbp32]{data/blur_learning.csv};
                
                \addplot[name path=deepbpd2,draw=none,fill=none] table[col sep=comma, x=x, y=deepbpd2]{data/blur_learning.csv};
                \addplot[name path=deepbpu2,draw=none,fill=none] table[col sep=comma, x=x, y=deepbpu2]{data/blur_learning.csv};
                \addplot[cdeepBP,fill opacity=0.15] fill between[of=deepbpd2 and deepbpu2];
                
                \addplot[name path=deepbpd8,draw=none,fill=none] table[col sep=comma, x=x, y=deepbpd8]{data/blur_learning.csv};
                \addplot[name path=deepbpu8,draw=none,fill=none] table[col sep=comma, x=x, y=deepbpu8]{data/blur_learning.csv};
                \addplot[cdeepBP8,fill opacity=0.15] fill between[of=deepbpd8 and deepbpu8];
                
                \addplot[name path=deepbpd32,draw=none,fill=none] table[col sep=comma, x=x, y=deepbpd32]{data/blur_learning.csv};
                \addplot[name path=deepbpu32,draw=none,fill=none] table[col sep=comma, x=x, y=deepbpu32]{data/blur_learning.csv};
                \addplot[cdeepBP32,fill opacity=0.15] fill between[of=deepbpd32 and deepbpu32];
                
                \addplot[name path=deepmfd2,draw=none,fill=none] table[col sep=comma, x=x, y=deepmfd2]{data/blur_learning.csv};
                \addplot[name path=deepmfu2,draw=none,fill=none] table[col sep=comma, x=x, y=deepmfu2]{data/blur_learning.csv};
                \addplot[cdeepMF,fill opacity=0.15] fill between[of=deepmfd2 and deepmfu2];
                
                \addplot[name path=deepmfd8,draw=none,fill=none] table[col sep=comma, x=x, y=deepmfd8]{data/blur_learning.csv};
                \addplot[name path=deepmfu8,draw=none,fill=none] table[col sep=comma, x=x, y=deepmfu8]{data/blur_learning.csv};
                \addplot[cdeepMF8,fill opacity=0.15] fill between[of=deepmfd8 and deepmfu8];
                
                \addplot[name path=deepmfd32,draw=none,fill=none] table[col sep=comma, x=x, y=deepmfd32]{data/blur_learning.csv};
                \addplot[name path=deepmfu32,draw=none,fill=none] table[col sep=comma, x=x, y=deepmfu32]{data/blur_learning.csv};
                \addplot[cdeepMF32,fill opacity=0.15] fill between[of=deepmfd32 and deepmfu32];
            \end{axis}
        \end{tikzpicture}
        \caption{Learning with blurred images}
        \label{fig:blur_learning}
    \end{subfigure}
    \caption{ {\em Robustness to blurred feature}. Inference (a) and learning (b)  performance on blurred images with clipping parameter $c$. $\text{deepMF}(r)$ and $\text{deepBP}(r)$ indicate the performance of algorithms on the dataset with $r$ radius of Gaussian blur. In all settings, the performance of deepMF and deepBP decrease as we blur images more. By clipping the classifier output with low values, we can bound the performance of deepMF and deepBP over the non-feature algorithms. We observe that deepBP outperforms the other existing methods.}
    \label{fig:feature_clip}
\end{figure}

\subsection{Robustness Analysis}
\label{sec:robustness}
Several inference and learning algorithms have been proposed, but their robustness on various environments has not been analyzed thoroughly yet. In this experiment, we test algorithmic robustness from different perspectives.
We use synthetic dataset for controlled experiments. The synthetic dataset is generated by assuming 1,000 tasks and 750 workers. The assignment structure between the tasks and workers is generated from a randomly sampled  $(l,r)$-regular bipartite graph, where $l$ indicates the number of workers assigned to each task and $r$ indicates the number of tasks per worker. Each task is a binary classification associated with an image feature sampled from the Dogs vs. Cats dataset~\citep{dogsvscats}. Workers' confusion matrices are randomly drawn from the one-coin model as \eqref{eq:one-coin-prior}, where we set the parameter of the Dirichlet distribution differently for each experiment.
We set $l=3$, $r=4$ for our synthetic data similar to the statistics of known datasets~\citep{snow2008cheap, welinder2010multidimensional, han2015faceage}. Note that the assignment graph satisfies the sparse regime reported in the earlier work~\citep{ok2016optimality}.
To evaluate the classifiers obtained from learning methods, we use the test set from the Dogs vs. Cats dataset~\citep{dogsvscats}.

\paragraph{Robustness to Feature.}
\label{sec:robustness to feature}
We first test robustness of inference and learning algorithms against non-informative features.
To generate a non-informative feature, we add Gaussian blur to the images\footnote{The examples of blurred images are in \Apxref{appendix:blur-images}} and vary the blur radius from $0$ to $32$. When features become more non-informative, i.e, as the radius increases, the message coming from task features becomes more irrelevant. To prevent the influence of features, we employ the clipping technique introduced in \Secref{sec:method-remark}. We vary the clipping parameter $c$ from $0.55$ to $1$.
The dataset is generated with worker prior $\text{Dir}(1, 0.5)$, and we use the true prior for experiments.

With informative features, \Figref{fig:blur_inference} shows the inference accuracy of deepMF and deepBP can be improved to compare with their non-feature counter-parts.
The learning accuracy is also increased with informative features as shown in \Figref{fig:blur_learning}.
The performance of feature-based algorithms decrease as the images get more corrupted.
However, we can still bound the performance of the proposed algorithms to those of non-feature algorithms by clipping the influence of the classifier when the features are highly non-informative.

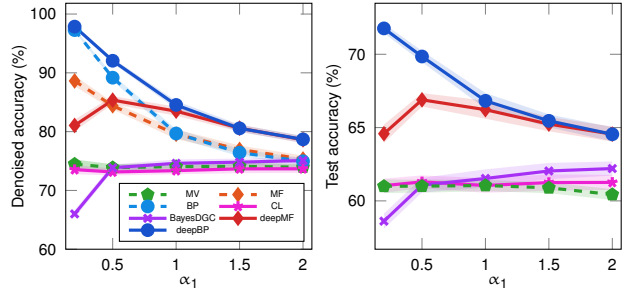
\begin{figure}[t!]
    \captionsetup[subfigure]{font=scriptsize,labelfont=scriptsize,aboveskip=0.05cm,belowskip=-0.15cm}
    \centering
    \begin{subfigure}[t]{.49\linewidth}
        \centering
        \begin{tikzpicture}
            \begin{axis}[
                legend style={nodes={scale=0.35, transform shape}}, 
                xlabel={$\alpha_1$},
                ylabel={Denoised accuracy (\%)},
                legend pos=south east,
                width=1.2\linewidth,
                height=1.2\linewidth,
                xlabel style={yshift=0.15cm},
                ylabel style={yshift=-0.2cm},
                legend columns=2,
                ymin=60,
                xmin=0.13,
                xmax=2.07
            ]
                \addplot[cMV, very thick, dashed, mark=pentagon*, mark size=2pt, mark options={solid}] table[col sep=comma, x=x, y=mv]{data/true_prior_inference.csv};
                \addplot[cMF, very thick, dashed, mark=diamond*, mark size=2pt, mark options={solid}] table[col sep=comma, x=x, y=mf]{data/true_prior_inference.csv};
                \addplot[cBP, very thick, dashed, mark=*, mark size=2pt, mark options={solid}] table[col sep=comma, x=x, y=bp]{data/true_prior_inference.csv};
                \addplot[cCL, very thick, mark=star, mark size=2pt, mark options={solid}] table[col sep=comma, x=x, y=cl]{data/true_prior_inference.csv};
                \addplot[cBayesDGC, very thick, mark=x, mark size=2pt, mark options={solid}] table[col sep=comma, x=x, y=bayesdgc]{data/true_prior_inference.csv};
                \addplot[cdeepMF, very thick, mark=diamond*, mark size=2pt, mark options={solid}] table[col sep=comma, x=x, y=deepmf]{data/true_prior_inference.csv};
                \addplot[cdeepBP, very thick, mark=*, mark size=2pt, mark options={solid}] table[col sep=comma, x=x, y=deepbp]{data/true_prior_inference.csv};
                
                \addplot[name path=deepbpd,draw=none,fill=none] table[col sep=comma, x=x, y=deepbpd]{data/true_prior_inference.csv};
                \addplot[name path=deepbpu,draw=none,fill=none] table[col sep=comma, x=x, y=deepbpu]{data/true_prior_inference.csv};
                \addplot[cdeepBP,fill opacity=0.15] fill between[of=deepbpd and deepbpu];
                
                \addplot[name path=bpd,draw=none,fill=none] table[col sep=comma, x=x, y=bpd]{data/true_prior_inference.csv};
                \addplot[name path=bpu,draw=none,fill=none] table[col sep=comma, x=x, y=bpu]{data/true_prior_inference.csv};
                \addplot[cBP,fill opacity=0.15] fill between[of=bpd and bpu];
                
                \addplot[name path=deepmfd,draw=none,fill=none] table[col sep=comma, x=x, y=deepmfd]{data/true_prior_inference.csv};
                \addplot[name path=deepmfu,draw=none,fill=none] table[col sep=comma, x=x, y=deepmfu]{data/true_prior_inference.csv};
                \addplot[cdeepMF,fill opacity=0.15] fill between[of=deepmfu and deepmfd];
                
                \addplot[name path=mfd,draw=none,fill=none] table[col sep=comma, x=x, y=mfd]{data/true_prior_inference.csv};
                \addplot[name path=mfu,draw=none,fill=none] table[col sep=comma, x=x, y=mfu]{data/true_prior_inference.csv};
                \addplot[cMF,fill opacity=0.15] fill between[of=mfu and mfd];
                
                \addplot[name path=mvd,draw=none,fill=none] table[col sep=comma, x=x, y=mvd]{data/true_prior_inference.csv};
                \addplot[name path=mvu,draw=none,fill=none] table[col sep=comma, x=x, y=mvu]{data/true_prior_inference.csv};
                \addplot[cMV,fill opacity=0.15] fill between[of=mvu and mvd];
                
                \addplot[name path=mvd,draw=none,fill=none] table[col sep=comma, x=x, y=mvd]{data/true_prior_inference.csv};
                \addplot[name path=mvu,draw=none,fill=none] table[col sep=comma, x=x, y=mvu]{data/true_prior_inference.csv};
                \addplot[cMV,fill opacity=0.15] fill between[of=mvu and mvd];
                
                \addplot[name path=cld,draw=none,fill=none] table[col sep=comma, x=x, y=cld]{data/true_prior_inference.csv};
                \addplot[name path=clu,draw=none,fill=none] table[col sep=comma, x=x, y=clu]{data/true_prior_inference.csv};
                \addplot[cCL,fill opacity=0.15] fill between[of=clu and cld];
                
                \addplot[name path=bayesdgcd,draw=none,fill=none] table[col sep=comma, x=x, y=bayesdgcd]{data/true_prior_inference.csv};
                \addplot[name path=bayesdgcu,draw=none,fill=none] table[col sep=comma, x=x, y=bayesdgcu]{data/true_prior_inference.csv};
                \addplot[cBayesDGC,fill opacity=0.15] fill between[of=bayesdgcu and bayesdgcd];
                
                \legend{MV,MF,BP,CL,BayesDGC,deepMF,deepBP}
        \end{axis}
        \end{tikzpicture}
        \caption{Inference with true prior}
        \label{fig:true_inference}
    \end{subfigure}
    \hspace{-1mm}
    \begin{subfigure}[t]{.49\linewidth}
        \centering
        \begin{tikzpicture}
            \begin{axis}[
                xlabel={$\alpha_1$},
                ylabel={Test accuracy (\%)},
                width=1.2\linewidth,
                height=1.2\linewidth,
                xlabel style={yshift=0.15cm},
                ylabel style={yshift=-0.15cm},
                xmin=0.13,
                xmax=2.07
            ]
                \addplot[cdeepMF, very thick, mark=diamond*, mark size=2pt, mark options={solid}] table[col sep=comma, x=x, y=deepmf]{data/true_prior_learning.csv};
                \addplot[cCL, very thick, mark=star, mark size=2pt, mark options={solid}] table[col sep=comma, x=x, y=cl]{data/true_prior_learning.csv};
                \addplot[cBayesDGC, very thick, mark=x, mark size=2pt, mark options={solid}] table[col sep=comma, x=x, y=bayesdgc]{data/true_prior_learning.csv};
                \addplot[cdeepBP, very thick, mark=*, mark size=2pt, mark options={solid}] table[col sep=comma, x=x, y=deepbp]{data/true_prior_learning.csv};
                \addplot[cMV, very thick, dashed, mark=pentagon*, mark size=2pt, mark options={solid}] table[col sep=comma, x=x, y=mv]{data/true_prior_learning.csv};
                
                \addplot[name path=deepbpd,draw=none,fill=none] table[col sep=comma, x=x, y=deepbpd]{data/true_prior_learning.csv};
                \addplot[name path=deepbpu,draw=none,fill=none] table[col sep=comma, x=x, y=deepbpu]{data/true_prior_learning.csv};
                \addplot[cdeepBP,fill opacity=0.15] fill between[of=deepbpd and deepbpu];
                
                \addplot[name path=deepmfd,draw=none,fill=none] table[col sep=comma, x=x, y=deepmfd]{data/true_prior_learning.csv};
                \addplot[name path=deepmfu,draw=none,fill=none] table[col sep=comma, x=x, y=deepmfu]{data/true_prior_learning.csv};
                \addplot[cdeepMF,fill opacity=0.15] fill between[of=deepmfu and deepmfd];
                
                \addplot[name path=cld,draw=none,fill=none] table[col sep=comma, x=x, y=cld]{data/true_prior_learning.csv};
                \addplot[name path=clu,draw=none,fill=none] table[col sep=comma, x=x, y=clu]{data/true_prior_learning.csv};
                \addplot[cCL,fill opacity=0.15] fill between[of=clu and cld];
                
                \addplot[name path=bayesdgcd,draw=none,fill=none] table[col sep=comma, x=x, y=bayesdgcd]{data/true_prior_learning.csv};
                \addplot[name path=bayesdgcu,draw=none,fill=none] table[col sep=comma, x=x, y=bayesdgcu]{data/true_prior_learning.csv};
                \addplot[cBayesDGC,fill opacity=0.15] fill between[of=bayesdgcu and bayesdgcd];
                
                \addplot[name path=mvd,draw=none,fill=none] table[col sep=comma, x=x, y=mvd]{data/true_prior_learning.csv};
                \addplot[name path=mvu,draw=none,fill=none] table[col sep=comma, x=x, y=mvu]{data/true_prior_learning.csv};
                \addplot[cMV,fill opacity=0.15] fill between[of=mvu and mvd];
            \end{axis}
        \end{tikzpicture}
        \caption{Learning with true prior}
        \label{fig:true_learning}
    \end{subfigure}
    
    \caption{ {\em Robustness to true prior.} Inference (a) and learning (b) performance with varying $\alpha_1$ of the true prior: $\text{Dir}(\alpha_1, \alpha_1 / 2)$.
    As the prior becomes sparser, i.e., $\alpha_1 < 1$, the gap between MF-based and BP-based algorithms increases. DeepBP outperforms the other methods across all settings.
    }
    \label{fig:true_prior}
\end{figure}

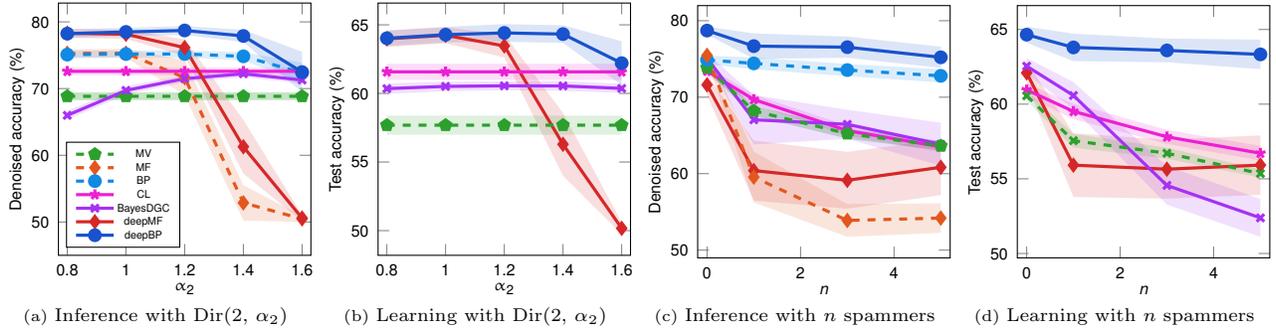
\begin{figure*}[!t]
    \captionsetup[subfigure]{font=scriptsize,labelfont=tiny,aboveskip=0.05cm,belowskip=-0.15cm}

    \centering
    \begin{subfigure}[t]{.24\linewidth}
        \centering
        \begin{tikzpicture}
            \begin{axis}[
                legend style={nodes={scale=0.4, transform shape}},
                legend pos=south west,
                xlabel={$\alpha_2$},
                ylabel={Denoised accuracy (\%)},
                legend pos=south west,
                width=1.2\linewidth,
                height=1.2\linewidth,
                xlabel style={yshift=0.15cm},
                ylabel style={yshift=-0.15cm},
                ymin=45,
                xmin=0.77,
                xmax=1.63
            ]
                \addplot[cMV, very thick, mark=pentagon*, dashed, mark size=2pt, mark options={solid}] table[col sep=comma, x=x, y=mv]{data/mismatched_extreme_spammer_inference.csv};
                \addplot[cMF, very thick, mark=diamond*, dashed, mark size=2pt, mark options={solid}] table[col sep=comma, x=x, y=mf]{data/mismatched_extreme_spammer_inference.csv};
                \addplot[cBP, very thick, mark=*, dashed, mark size=2pt, mark options={solid}] table[col sep=comma, x=x, y=bp]{data/mismatched_extreme_spammer_inference.csv};
                \addplot[cCL, very thick, mark=star, , mark size=2pt, mark options={solid}] table[col sep=comma, x=x, y=cl]{data/mismatched_extreme_spammer_inference.csv};
                \addplot[cBayesDGC, very thick, mark=x, mark size=2pt, mark options={solid}] table[col sep=comma, x=x, y=bayesdgc]{data/mismatched_extreme_spammer_inference.csv};
                \addplot[cdeepMF, very thick, mark=diamond*, mark size=2pt, mark options={solid}] table[col sep=comma, x=x, y=deepmf]{data/mismatched_extreme_spammer_inference.csv};
                \addplot[cdeepBP, very thick, mark=*, mark size=2pt, mark options={solid}] table[col sep=comma, x=x, y=deepbp]{data/mismatched_extreme_spammer_inference.csv};   
                
                \addplot[name path=deepbpd,draw=none,fill=none] table[col sep=comma, x=x, y=deepbpd]{data/mismatched_extreme_spammer_inference.csv};
                \addplot[name path=deepbpu,draw=none,fill=none] table[col sep=comma, x=x, y=deepbpu]{data/mismatched_extreme_spammer_inference.csv};
                \addplot[cdeepBP,fill opacity=0.15] fill between[of=deepbpd and deepbpu];
                
                \addplot[name path=bpd,draw=none,fill=none] table[col sep=comma, x=x, y=bpd]{data/mismatched_extreme_spammer_inference.csv};
                \addplot[name path=bpu,draw=none,fill=none] table[col sep=comma, x=x, y=bpu]{data/mismatched_extreme_spammer_inference.csv};
                \addplot[cBP,fill opacity=0.15] fill between[of=bpd and bpu];
                
                \addplot[name path=deepmfd,draw=none,fill=none] table[col sep=comma, x=x, y=deepmfd]{data/mismatched_extreme_spammer_inference.csv};
                \addplot[name path=deepmfu,draw=none,fill=none] table[col sep=comma, x=x, y=deepmfu]{data/mismatched_extreme_spammer_inference.csv};
                \addplot[cdeepMF,fill opacity=0.15] fill between[of=deepmfu and deepmfd];
                
                \addplot[name path=mfd,draw=none,fill=none] table[col sep=comma, x=x, y=mfd]{data/mismatched_extreme_spammer_inference.csv};
                \addplot[name path=mfu,draw=none,fill=none] table[col sep=comma, x=x, y=mfu]{data/mismatched_extreme_spammer_inference.csv};
                \addplot[cMF,fill opacity=0.15] fill between[of=mfu and mfd];
                
                \addplot[name path=mvd,draw=none,fill=none] table[col sep=comma, x=x, y=mvd]{data/mismatched_extreme_spammer_inference.csv};
                \addplot[name path=mvu,draw=none,fill=none] table[col sep=comma, x=x, y=mvu]{data/mismatched_extreme_spammer_inference.csv};
                \addplot[cMV,fill opacity=0.15] fill between[of=mvu and mvd];
                
                \addplot[name path=mvd,draw=none,fill=none] table[col sep=comma, x=x, y=mvd]{data/mismatched_extreme_spammer_inference.csv};
                \addplot[name path=mvu,draw=none,fill=none] table[col sep=comma, x=x, y=mvu]{data/mismatched_extreme_spammer_inference.csv};
                \addplot[cMV,fill opacity=0.15] fill between[of=mvu and mvd];
                
                \addplot[name path=cld,draw=none,fill=none] table[col sep=comma, x=x, y=cld]{data/mismatched_extreme_spammer_inference.csv};
                \addplot[name path=clu,draw=none,fill=none] table[col sep=comma, x=x, y=clu]{data/mismatched_extreme_spammer_inference.csv};
                \addplot[cCL,fill opacity=0.15] fill between[of=clu and cld];
                
                \addplot[name path=bayesdgcd,draw=none,fill=none] table[col sep=comma, x=x, y=bayesdgcd]{data/mismatched_extreme_spammer_inference.csv};
                \addplot[name path=bayesdgcu,draw=none,fill=none] table[col sep=comma, x=x, y=bayesdgcu]{data/mismatched_extreme_spammer_inference.csv};
                \addplot[cBayesDGC,fill opacity=0.15] fill between[of=bayesdgcu and bayesdgcd];
                
                \legend{MV,MF,BP,CL,BayesDGC,deepMF,deepBP}
            \end{axis}
        \end{tikzpicture}
        \caption{Inference with Dir(2, $\alpha_2$)}
        \label{fig:extreme_spammer_mismatched_inference}
    \end{subfigure} \hspace{-1mm} %
    \begin{subfigure}[t]{.24\linewidth}
        \centering
        \begin{tikzpicture}
            \begin{axis}[
                xlabel={$\alpha_2$},
                ylabel={Test accuracy (\%)},
                width=1.2\linewidth,
                height=1.2\linewidth,
                xlabel style={yshift=0.15cm},
                ylabel style={yshift=-0.15cm},
                xmin=0.77,
                xmax=1.63
            ]
                \addplot[cdeepMF, very thick, mark=diamond*, mark size=2pt, mark options={solid}] table[col sep=comma, x=x, y=deepmf]{data/mismatched_extreme_spammer_learning.csv};
                \addplot[cCL, very thick, mark=star, mark size=2pt, mark options={solid}] table[col sep=comma, x=x, y=cl]{data/mismatched_extreme_spammer_learning.csv};
                \addplot[cBayesDGC, very thick, mark=x, mark size=2pt, mark options={solid}] table[col sep=comma, x=x, y=bayesdgc]{data/mismatched_extreme_spammer_learning.csv};
                \addplot[cdeepBP, very thick, mark=*, mark size=2pt, mark options={solid}] table[col sep=comma, x=x, y=deepbp]{data/mismatched_extreme_spammer_learning.csv};
                \addplot[cMV, very thick, mark=pentagon*, dashed, mark size=2pt, mark options={solid}] table[col sep=comma, x=x, y=mv]{data/mismatched_extreme_spammer_learning.csv};
                
                \addplot[name path=deepbpd,draw=none,fill=none] table[col sep=comma, x=x, y=deepbpd]{data/mismatched_extreme_spammer_learning.csv};
                \addplot[name path=deepbpu,draw=none,fill=none] table[col sep=comma, x=x, y=deepbpu]{data/mismatched_extreme_spammer_learning.csv};
                \addplot[cdeepBP,fill opacity=0.15] fill between[of=deepbpd and deepbpu];
                
                \addplot[name path=deepmfd,draw=none,fill=none] table[col sep=comma, x=x, y=deepmfd]{data/mismatched_extreme_spammer_learning.csv};
                \addplot[name path=deepmfu,draw=none,fill=none] table[col sep=comma, x=x, y=deepmfu]{data/mismatched_extreme_spammer_learning.csv};
                \addplot[cdeepMF,fill opacity=0.15] fill between[of=deepmfu and deepmfd];
                
                \addplot[name path=cld,draw=none,fill=none] table[col sep=comma, x=x, y=cld]{data/mismatched_extreme_spammer_learning.csv};
                \addplot[name path=clu,draw=none,fill=none] table[col sep=comma, x=x, y=clu]{data/mismatched_extreme_spammer_learning.csv};
                \addplot[cCL,fill opacity=0.15] fill between[of=clu and cld];
                
                \addplot[name path=bayesdgcd,draw=none,fill=none] table[col sep=comma, x=x, y=bayesdgcd]{data/mismatched_extreme_spammer_learning.csv};
                \addplot[name path=bayesdgcu,draw=none,fill=none] table[col sep=comma, x=x, y=bayesdgcu]{data/mismatched_extreme_spammer_learning.csv};
                \addplot[cBayesDGC,fill opacity=0.15] fill between[of=bayesdgcu and bayesdgcd];
                
                \addplot[name path=mvd,draw=none,fill=none] table[col sep=comma, x=x, y=mvd]{data/mismatched_extreme_spammer_learning.csv};
                \addplot[name path=mvu,draw=none,fill=none] table[col sep=comma, x=x, y=mvu]{data/mismatched_extreme_spammer_learning.csv};
                \addplot[cMV,fill opacity=0.15] fill between[of=mvu and mvd];
            \end{axis}
        \end{tikzpicture}
        \caption{Learning with Dir(2, $\alpha_2$)}
        \label{fig:extreme_spammer_mismatched_learning}
    \end{subfigure} \hspace{-1mm} %
    \begin{subfigure}[t]{.24\linewidth}
        \centering
        \begin{tikzpicture}
            \begin{axis}[
                xlabel={$n$},
                ylabel={Denoised accuracy (\%)},
                width=1.2\linewidth,
                height=1.2\linewidth,
                xlabel style={yshift=0.15cm},
                ylabel style={yshift=-0.15cm},
                xmin=-0.2,
                xmax=5.2
            ]
                \addplot[cdeepBP, very thick, mark=*, mark size=2pt, mark options={solid}] table[col sep=comma, x=x, y=deepbp]{data/num_spam_inference.csv};
                \addplot[cdeepMF, very thick, mark=diamond*, mark size=2pt, mark options={solid}] table[col sep=comma, x=x, y=deepmf]{data/num_spam_inference.csv};
                \addplot[cCL, very thick, mark=star, mark size=2pt, mark options={solid}] table[col sep=comma, x=x, y=cl]{data/num_spam_inference.csv};
                \addplot[cBayesDGC, very thick, mark=x, mark size=2pt, mark options={solid}] table[col sep=comma, x=x, y=bayesdgc]{data/num_spam_inference.csv};
                \addplot[cMV, very thick, mark=pentagon*, dashed, mark size=2pt, mark options={solid}] table[col sep=comma, x=x, y=mv]{data/num_spam_inference.csv};
                \addplot[cBP, very thick, mark=*, dashed, mark size=2pt, mark options={solid}] table[col sep=comma, x=x, y=bp]{data/num_spam_inference.csv};
                \addplot[cMF, very thick, mark=diamond*, dashed, mark size=2pt, mark options={solid}] table[col sep=comma, x=x, y=mf]{data/num_spam_inference.csv};
                
                \addplot[name path=deepbpd,draw=none,fill=none] table[col sep=comma, x=x, y=deepbpd]{data/num_spam_inference.csv};
                \addplot[name path=deepbpu,draw=none,fill=none] table[col sep=comma, x=x, y=deepbpu]{data/num_spam_inference.csv};
                \addplot[cdeepBP,fill opacity=0.15] fill between[of=deepbpd and deepbpu];
                
                \addplot[name path=bpd,draw=none,fill=none] table[col sep=comma, x=x, y=bpd]{data/num_spam_inference.csv};
                \addplot[name path=bpu,draw=none,fill=none] table[col sep=comma, x=x, y=bpu]{data/num_spam_inference.csv};
                \addplot[cBP,fill opacity=0.15] fill between[of=bpd and bpu];
                
                \addplot[name path=deepmfd,draw=none,fill=none] table[col sep=comma, x=x, y=deepmfd]{data/num_spam_inference.csv};
                \addplot[name path=deepmfu,draw=none,fill=none] table[col sep=comma, x=x, y=deepmfu]{data/num_spam_inference.csv};
                \addplot[cdeepMF,fill opacity=0.15] fill between[of=deepmfu and deepmfd];
                
                \addplot[name path=mfd,draw=none,fill=none] table[col sep=comma, x=x, y=mfd]{data/num_spam_inference.csv};
                \addplot[name path=mfu,draw=none,fill=none] table[col sep=comma, x=x, y=mfu]{data/num_spam_inference.csv};
                \addplot[cMF,fill opacity=0.15] fill between[of=mfu and mfd];
                
                \addplot[name path=mvd,draw=none,fill=none] table[col sep=comma, x=x, y=mvd]{data/num_spam_inference.csv};
                \addplot[name path=mvu,draw=none,fill=none] table[col sep=comma, x=x, y=mvu]{data/num_spam_inference.csv};
                \addplot[cMV,fill opacity=0.15] fill between[of=mvu and mvd];
                
                \addplot[name path=mvd,draw=none,fill=none] table[col sep=comma, x=x, y=mvd]{data/num_spam_inference.csv};
                \addplot[name path=mvu,draw=none,fill=none] table[col sep=comma, x=x, y=mvu]{data/num_spam_inference.csv};
                \addplot[cMV,fill opacity=0.15] fill between[of=mvu and mvd];
                
                \addplot[name path=cld,draw=none,fill=none] table[col sep=comma, x=x, y=cld]{data/num_spam_inference.csv};
                \addplot[name path=clu,draw=none,fill=none] table[col sep=comma, x=x, y=clu]{data/num_spam_inference.csv};
                \addplot[cCL,fill opacity=0.15] fill between[of=clu and cld];
                
                \addplot[name path=bayesdgcd,draw=none,fill=none] table[col sep=comma, x=x, y=bayesdgcd]{data/num_spam_inference.csv};
                \addplot[name path=bayesdgcu,draw=none,fill=none] table[col sep=comma, x=x, y=bayesdgcu]{data/num_spam_inference.csv};
                \addplot[cBayesDGC,fill opacity=0.15] fill between[of=bayesdgcu and bayesdgcd];
                
            \end{axis}
        \end{tikzpicture}
        \caption{Inference with $n$ spammers}
        \label{fig:extreme_spammer_frequent_inference}
    \end{subfigure} \hspace{-1mm} %
    \begin{subfigure}[t]{.24\linewidth}
        \centering
        \begin{tikzpicture}
            \begin{axis}[
                xlabel={$n$},
                ylabel={Test accuracy (\%)},
                width=1.2\linewidth,
                height=1.2\linewidth,
                xlabel style={yshift=0.15cm},
                ylabel style={yshift=-0.15cm},
                xmin=-0.2,
                xmax=5.2
            ]
                \addplot[cdeepBP, very thick, mark=*, mark size=2pt, mark options={solid}] table[col sep=comma, x=x, y=deepbp]{data/num_spam_learning.csv};
                \addplot[cdeepMF, very thick, mark=diamond*, mark size=2pt, mark options={solid}] table[col sep=comma, x=x, y=deepmf]{data/num_spam_learning.csv};
                \addplot[cCL, very thick, mark=star, mark size=2pt, mark options={solid}] table[col sep=comma, x=x, y=cl]{data/num_spam_learning.csv};
                \addplot[cBayesDGC, very thick, mark=x, mark size=2pt, mark options={solid}] table[col sep=comma, x=x, y=bayesdgc]{data/num_spam_learning.csv};
                \addplot[cMV, very thick, mark=pentagon*, dashed, mark=x, mark size=2pt, mark options={solid}] table[col sep=comma, x=x, y=mv]{data/num_spam_learning.csv};
                
                \addplot[name path=deepbpd,draw=none,fill=none] table[col sep=comma, x=x, y=deepbpd]{data/num_spam_learning.csv};
                \addplot[name path=deepbpu,draw=none,fill=none] table[col sep=comma, x=x, y=deepbpu]{data/num_spam_learning.csv};
                \addplot[cdeepBP,fill opacity=0.15] fill between[of=deepbpd and deepbpu];
                
                \addplot[name path=deepmfd,draw=none,fill=none] table[col sep=comma, x=x, y=deepmfd]{data/num_spam_learning.csv};
                \addplot[name path=deepmfu,draw=none,fill=none] table[col sep=comma, x=x, y=deepmfu]{data/num_spam_learning.csv};
                \addplot[cdeepMF,fill opacity=0.15] fill between[of=deepmfu and deepmfd];
                
                \addplot[name path=cld,draw=none,fill=none] table[col sep=comma, x=x, y=cld]{data/num_spam_learning.csv};
                \addplot[name path=clu,draw=none,fill=none] table[col sep=comma, x=x, y=clu]{data/num_spam_learning.csv};
                \addplot[cCL,fill opacity=0.15] fill between[of=clu and cld];
                
                \addplot[name path=bayesdgcd,draw=none,fill=none] table[col sep=comma, x=x, y=bayesdgcd]{data/num_spam_learning.csv};
                \addplot[name path=bayesdgcu,draw=none,fill=none] table[col sep=comma, x=x, y=bayesdgcu]{data/num_spam_learning.csv};
                \addplot[cBayesDGC,fill opacity=0.15] fill between[of=bayesdgcu and bayesdgcd];
                
                \addplot[name path=mvd,draw=none,fill=none] table[col sep=comma, x=x, y=mvd]{data/num_spam_learning.csv};
                \addplot[name path=mvu,draw=none,fill=none] table[col sep=comma, x=x, y=mvu]{data/num_spam_learning.csv};
                \addplot[cMV,fill opacity=0.15] fill between[of=mvu and mvd];
            \end{axis}
        \end{tikzpicture}
        \caption{Learning with $n$ spammers}
        \label{fig:extreme_spammer_frequent_learning}
    \end{subfigure}
    \caption{
    {\em Robustness to extreme-spammers with mismatched prior.} (a, b) The performance of MF-based methods decrease as the model prior deviates from the true prior: Dir$(2,1)$.  BP-based methods perform robustly than the MF-based methods in both inference and learning. (c, d) BP-based methods are influenced less from the spammers than the other methods regardless of the number of spammers. We use Dir(2,1.4) for the model prior.}
    \label{fig:extreme_spammer}
\end{figure*}

\begin{figure*}[t]
    \captionsetup[subfigure]{font=scriptsize,labelfont=scriptsize,aboveskip=0.05cm,belowskip=-0.15cm}
    \centering
    \begin{subfigure}[t]{.32\linewidth}
        \centering
        \begin{tikzpicture}
            \begin{axis}[
                legend style={nodes={scale=0.5, transform shape}}, 
                area style,
                width=0.85\linewidth,
                scale only axis, 
                height=0.17\linewidth, 
                ymin=0,
                legend pos=north west,
                ymax=200,
                ylabel={Number of tasks},
                ylabel style={xshift=-0.5cm, yshift=-0.15cm}
            ]
                \addplot+[hist={bins=10, data max=1, data min=0}, fill=cBP, draw=cBP!60!black] table[col sep=comma, y=bp]{data/histogram.csv};
                \legend{BP};
            \end{axis}
            
            \begin{axis}[
                legend style={nodes={scale=0.5, transform shape}}, 
                area style,
                width=0.85\linewidth, 
                xlabel={$q(z=0)$},
                scale only axis, 
                height=0.17\linewidth, 
                ymin=0,
                yshift=-0.24\linewidth,
                legend pos=north west,
                ymax=200,
                xlabel style={yshift=0.15cm}
            ]
                \addplot+[hist={bins=10, data max=1, data min=0}, fill=cMF, draw=cMF!60!black] table[col sep=comma, y=mf]{data/histogram.csv};
                \legend{MF};
            \end{axis}
        \end{tikzpicture}
        \caption{Histogram of the marginals}
        \label{fig:marginal}
    \end{subfigure}
    \hspace{-1mm}
    \begin{subfigure}[t]{.32\linewidth}
        \centering
        \begin{tikzpicture}
            \begin{axis}[
                legend style={nodes={scale=0.5, transform shape}}, 
                ylabel={Denoised Accuracy (\%)}, 
                xlabel={Portion of worst results (\%)},
                width=0.85\linewidth,
                height=0.43\linewidth,
                scale only axis, 
                legend pos=south east,
                ylabel style={yshift=-0.15cm},
                xlabel style={yshift=0.15cm}
            ]
                \addplot[cBP, very thick, mark=*, dashed, mark size=2pt, mark options={solid}] table[col sep=comma, x=x, y=bp]{data/cvar_metric_feature.csv};
                \addplot[cMF, very thick, mark=diamond*, dashed, mark size=2pt, mark options={solid}] table[col sep=comma, x=x, y=mf]{data/cvar_metric_feature.csv};
                \legend{BP,MF};
            \end{axis}
        \end{tikzpicture}
        \caption{Average performance of the worst results}
        \label{fig:cvar}
    \end{subfigure}
    \hspace{-1mm}
    \begin{subfigure}[t]{.32\linewidth}
        \centering
        \begin{tikzpicture}
            \begin{axis}[
                area style,
                xlabel={$q(\mTheta_{kk})$},
                width=.85\linewidth,
                scale only axis, 
                height=.43\linewidth,
                ymode=log,
                xmax=1.1,
                ylabel=Number of diagonal elements,
                ylabel style={yshift=-0.15cm},
                xlabel style={yshift=0.15cm}
            ]
                \addplot+[hist={bins=1, data min=0, data max=0.1}, thick, fill=cMF, draw=black, postaction={pattern=north east lines}] table[y index=0]{data/extreme_spammer_abilities.csv};
                \addplot+[hist={bins=8, data min=0.2, data max=0.9}, fill=cMF, draw=cMF!60!black] table[y index=0]{data/worker_abilities.csv};
                \draw [->, thick] (0.035\linewidth, 0.06\linewidth)--++(0, 0.19\linewidth);
                \draw (0.1\linewidth, 0.35\linewidth) node {\tiny\textsf{extreme-spammer}};
            \end{axis}
        \end{tikzpicture}
        \caption{Histogram of $q(\theta_{kk})$ when MF fails}
        \label{fig:extreme_spammer_overconfidence}
    \end{subfigure}
    \caption{{\em Overconfidence of MF-based methods.} 
    (a) The distribution of marginals on $\rvz$ obtained from the variational distribution $q(\rvz)$. 
    (b) The inference accuracy sorted in increasing order from multiple experiments with different seeds. In the worst case, the performance of MF significantly worse than that of BP, although their performances are not on average. (c) Histogram of diagonal elements in confusion matrices when MF fails. MF is likely to categorize the extreme-spammer as an adversary or a hammer.}
    \label{fig:overconfidence}
\end{figure*}
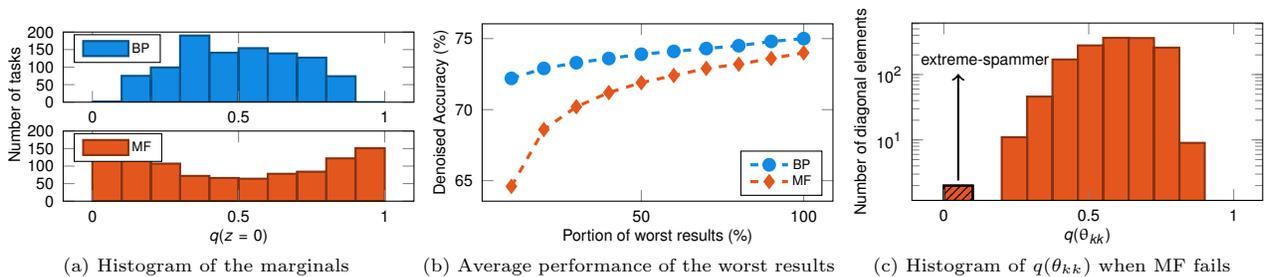

\paragraph{Robustness to Prior.}
A prior distribution over the confusion matrices needs to be set a priori. The exact prior is unknown in general, but in the following experiments, we investigate how the algorithms work when the models know the true prior.
This study reveals a failure of algorithms even with the exact knowledge on the true prior.
We use Dir$(\alpha_1, \alpha_1 / 2)$ with varying $\alpha_1$ from $0.2$ to $2$ as a prior over the confusion matrices.
As $\alpha_1$ goes to zero, the prior becomes sparse, i.e., the worker is either an adversary or a hammer. The population of spammer increases as $\alpha_1$ goes to two.

The results on the inference and learning tasks in \Figref{fig:true_prior} show the MF-based algorithms perform worse than BP-based when the prior distribution is sparse, i.e., $\alpha_1 < 1$. Although deepMF performs better than MF except the sparse prior case, both algorithms perform worse than their BP counter-parts.

\paragraph{Robustness to Extreme-spammer.}
\label{sec:extreme-spammer}
In a typical crowdsourcing system, workers get a fixed amount of reward by solving a single task, and therefore some workers attempt as many tasks as possible to maximize the reward~\citep{gadiraju2015human}. Among those, we focus on \textit{extreme-spammer}, who labels uniformly across all tasks, i.e. $p(y_i^{(u)}=z_i) \approx 1/K$.
Indeed, we observe the extreme-spammer appears frequently in real-world datasets. 
Check
\Apxref{appendix:dataset-details} 
for the existence of the extreme-spammer.
We add the extreme-spammer to a synthetic dataset with worker prior $\text{Dir}(2, 1)$ to see the effect of an extreme-spammer on each algorithm. We further assume that the true prior is unknown, making the simulated environment more realistic. All algorithms are evaluated with $\text{Dir}(2, \alpha_2)$, where $\alpha_2$ varies from $0.8$ to $1.6$.

\Figref{fig:extreme_spammer_mismatched_inference} and \Figref{fig:extreme_spammer_mismatched_learning} show the inference and learning accuracy with an extreme-spammer, respectively.
When the true prior is given, i.e. $\alpha_2 = 1$, both MF-based and BP-based algorithms perform well.
However, the accuracy of MF-based algorithms starts to decrease as the mismatch between true and model prior becomes bigger. BP-based algorithms are robust to extreme-spammer than MF-based algorithms regardless of the given prior in this setting.
When we add more extreme-spammers, the accuracy of BP-based algorithms decreases less than those of MF-based algorithms. The results are shown in \Figref{fig:extreme_spammer_frequent_inference} and \Figref{fig:extreme_spammer_frequent_learning}. This result suggests that MF-based algorithms are more vulnerable to a few workers who labels a lot. 

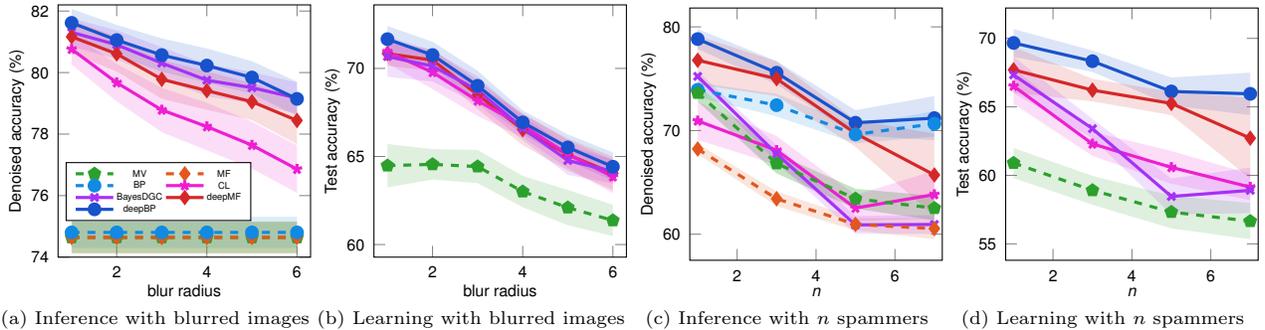
\begin{figure*}[th!]
    \captionsetup[subfigure]{font=scriptsize,labelfont=scriptsize,aboveskip=0.05cm,belowskip=-0.15cm}
    \centering
    \begin{subfigure}[t]{.24\linewidth}
        \centering
        \begin{tikzpicture}
            \begin{axis}[
                legend style={nodes={scale=0.35}, at={(0.03, 0.26)}, anchor=west}, 
                xlabel={blur radius},
                ylabel={Denoised accuracy (\%)},
                width=1.2\linewidth,
                height=1.2\linewidth,
                xlabel style={yshift=0.15cm},
                ylabel style={yshift=-0.15cm},
                xmin=0.7,
                xmax=6.3,
                ymin=74,
                ymax=82.2,
                legend columns=2
            ]
                \addplot[cMV, very thick, mark=pentagon*, dashed, mark size=2pt, mark options={solid}] table[col sep=comma, x=x, y=mv]{data/real_world_blur_inference.csv};
                \addplot[cMF, very thick, mark=diamond*, dashed, mark size=2pt, mark options={solid}] table[col sep=comma, x=x, y=mf]{data/real_world_blur_inference.csv};
                \addplot[cBP, very thick, mark=*, dashed, mark size=2pt, mark options={solid}] table[col sep=comma, x=x, y=bp]{data/real_world_blur_inference.csv};
                \addplot[cCL, very thick, mark=star, mark size=2pt, mark options={solid}] table[col sep=comma, x=x, y=cl]{data/real_world_blur_inference.csv};
                \addplot[cBayesDGC, very thick, mark=x, mark size=2pt, mark options={solid}] table[col sep=comma, x=x, y=bayesdgc]{data/real_world_blur_inference.csv};
                \addplot[cdeepMF, very thick, mark=diamond*, mark size=2pt, mark options={solid}] table[col sep=comma, x=x, y=deepmf]{data/real_world_blur_inference.csv};
                \addplot[cdeepBP, very thick, mark=*, mark size=2pt, mark options={solid}] table[col sep=comma, x=x, y=deepbp]{data/real_world_blur_inference.csv};
                \legend{MV,MF,BP,CL,BayesDGC,deepMF,deepBP}
                
                \addplot[name path=deepbpd,draw=none,fill=none] table[col sep=comma, x=x, y=deepbpd]{data/real_world_blur_inference.csv};
                \addplot[name path=deepbpu,draw=none,fill=none] table[col sep=comma, x=x, y=deepbpu]{data/real_world_blur_inference.csv};
                \addplot[cdeepBP,fill opacity=0.15] fill between[of=deepbpd and deepbpu];
                
                \addplot[name path=bpd,draw=none,fill=none] table[col sep=comma, x=x, y=bpd]{data/real_world_blur_inference.csv};
                \addplot[name path=bpu,draw=none,fill=none] table[col sep=comma, x=x, y=bpu]{data/real_world_blur_inference.csv};
                \addplot[cBP,fill opacity=0.15] fill between[of=bpd and bpu];
                
                \addplot[name path=deepmfd,draw=none,fill=none] table[col sep=comma, x=x, y=deepmfd]{data/real_world_blur_inference.csv};
                \addplot[name path=deepmfu,draw=none,fill=none] table[col sep=comma, x=x, y=deepmfu]{data/real_world_blur_inference.csv};
                \addplot[cdeepMF,fill opacity=0.15] fill between[of=deepmfu and deepmfd];
                
                \addplot[name path=mfd,draw=none,fill=none] table[col sep=comma, x=x, y=mfd]{data/real_world_blur_inference.csv};
                \addplot[name path=mfu,draw=none,fill=none] table[col sep=comma, x=x, y=mfu]{data/real_world_blur_inference.csv};
                \addplot[cMF,fill opacity=0.15] fill between[of=mfu and mfd];
                
                \addplot[name path=mvd,draw=none,fill=none] table[col sep=comma, x=x, y=mvd]{data/real_world_blur_inference.csv};
                \addplot[name path=mvu,draw=none,fill=none] table[col sep=comma, x=x, y=mvu]{data/real_world_blur_inference.csv};
                \addplot[cMV,fill opacity=0.15] fill between[of=mvu and mvd];
                
                \addplot[name path=mvd,draw=none,fill=none] table[col sep=comma, x=x, y=mvd]{data/real_world_blur_inference.csv};
                \addplot[name path=mvu,draw=none,fill=none] table[col sep=comma, x=x, y=mvu]{data/real_world_blur_inference.csv};
                \addplot[cMV,fill opacity=0.15] fill between[of=mvu and mvd];
                
                \addplot[name path=cld,draw=none,fill=none] table[col sep=comma, x=x, y=cld]{data/real_world_blur_inference.csv};
                \addplot[name path=clu,draw=none,fill=none] table[col sep=comma, x=x, y=clu]{data/real_world_blur_inference.csv};
                \addplot[cCL,fill opacity=0.15] fill between[of=clu and cld];
                
                \addplot[name path=bayesdgcd,draw=none,fill=none] table[col sep=comma, x=x, y=bayesdgcd]{data/real_world_blur_inference.csv};
                \addplot[name path=bayesdgcu,draw=none,fill=none] table[col sep=comma, x=x, y=bayesdgcu]{data/real_world_blur_inference.csv};
                \addplot[cBayesDGC,fill opacity=0.15] fill between[of=bayesdgcu and bayesdgcd];    
                
            \end{axis}
        \end{tikzpicture}
        \caption{Inference with blurred images}
        \label{fig:real_world_blur_inference}
    \end{subfigure} \hspace{-1.5mm} %
    \begin{subfigure}[t]{.24\linewidth}
        \centering
        \begin{tikzpicture}
            \begin{axis}[
                xlabel={blur radius},
                ylabel={Test accuracy (\%)},
                width=1.2\linewidth,
                height=1.2\linewidth,
                xlabel style={yshift=0.15cm},
                ylabel style={yshift=-0.15cm},
                xmin=0.7,
                xmax=6.3
            ]
                \addplot[cdeepMF, very thick, mark=diamond*, mark size=2pt, mark options={solid}] table[col sep=comma, x=x, y=deepmf]{data/real_world_blur_learning.csv};
                \addplot[cCL, very thick, mark=star, mark size=2pt, mark options={solid}] table[col sep=comma, x=x, y=cl]{data/real_world_blur_learning.csv};
                \addplot[cBayesDGC, very thick, mark=x, mark size=2pt, mark options={solid}] table[col sep=comma, x=x, y=bayesdgc]{data/real_world_blur_learning.csv};
                \addplot[cdeepBP, very thick, mark=*, mark size=2pt, mark options={solid}] table[col sep=comma, x=x, y=deepbp]{data/real_world_blur_learning.csv};
                \addplot[cMV, very thick, mark=pentagon*, dashed, mark size=2pt, mark options={solid}] table[col sep=comma, x=x, y=mv]{data/real_world_blur_learning.csv};
                
                \addplot[name path=deepbpd,draw=none,fill=none] table[col sep=comma, x=x, y=deepbpd]{data/real_world_blur_learning.csv};
                \addplot[name path=deepbpu,draw=none,fill=none] table[col sep=comma, x=x, y=deepbpu]{data/real_world_blur_learning.csv};
                \addplot[cdeepBP,fill opacity=0.15] fill between[of=deepbpd and deepbpu];
                
                \addplot[name path=deepmfd,draw=none,fill=none] table[col sep=comma, x=x, y=deepmfd]{data/real_world_blur_learning.csv};
                \addplot[name path=deepmfu,draw=none,fill=none] table[col sep=comma, x=x, y=deepmfu]{data/real_world_blur_learning.csv};
                \addplot[cdeepMF,fill opacity=0.15] fill between[of=deepmfu and deepmfd];
                
                \addplot[name path=cld,draw=none,fill=none] table[col sep=comma, x=x, y=clda]{data/real_world_blur_learning.csv};
                \addplot[name path=clu,draw=none,fill=none] table[col sep=comma, x=x, y=clu]{data/real_world_blur_learning.csv};
                \addplot[cCL,fill opacity=0.15] fill between[of=clu and cld];
                
                \addplot[name path=bayesdgcd,draw=none,fill=none] table[col sep=comma, x=x, y=bayesdgcd]{data/real_world_blur_learning.csv};
                \addplot[name path=bayesdgcu,draw=none,fill=none] table[col sep=comma, x=x, y=bayesdgcu]{data/real_world_blur_learning.csv};
                \addplot[cBayesDGC,fill opacity=0.15] fill between[of=bayesdgcu and bayesdgcd];
                
                \addplot[name path=mvd,draw=none,fill=none] table[col sep=comma, x=x, y=mvd]{data/real_world_blur_learning.csv};
                \addplot[name path=mvu,draw=none,fill=none] table[col sep=comma, x=x, y=mvu]{data/real_world_blur_learning.csv};
                \addplot[cMV,fill opacity=0.15] fill between[of=mvu and mvd];
            \end{axis}
        \end{tikzpicture}
        \caption{Learning with blurred images}
        \label{fig:real_world_blur_learning}
    \end{subfigure} \hspace{-1.5mm} %
    \begin{subfigure}[t]{.24\linewidth}
        \centering
        \begin{tikzpicture}
            \begin{axis}[
                xlabel={$n$},
                ylabel={Denoised accuracy (\%)},
                width=1.2\linewidth,
                height=1.2\linewidth,
                xlabel style={yshift=0.15cm},
                ylabel style={yshift=-0.15cm},
                xmax=7.2,
                xmin=0.8
            ]
                \addplot[cdeepBP, very thick, mark=*, mark size=2pt, mark options={solid}] table[col sep=comma, x=x, y=deepbp]{data/real_inference.csv};
                \addplot[cdeepMF, very thick, mark=diamond*, mark size=2pt, mark options={solid}] table[col sep=comma, x=x, y=deepmf]{data/real_inference.csv};
                \addplot[cCL, very thick, mark=star, mark size=2pt, mark options={solid}] table[col sep=comma, x=x, y=cl]{data/real_inference.csv};
                \addplot[cBayesDGC, very thick, mark=x, mark size=2pt, mark options={solid}] table[col sep=comma, x=x, y=bayesdgc]{data/real_inference.csv};
                \addplot[cBP, very thick, mark=*, dashed, mark size=2pt, mark options={solid}] table[col sep=comma, x=x, y=bp]{data/real_inference.csv};
                \addplot[cMF, very thick, mark=diamond*, dashed, mark size=2pt, mark options={solid}] table[col sep=comma, x=x, y=mf]{data/real_inference.csv};
                \addplot[cMV, very thick, mark=pentagon*, dashed, mark size=2pt, mark options={solid}] table[col sep=comma, x=x, y=mv]{data/real_inference.csv};
                
                \addplot[name path=deepbpd,draw=none,fill=none] table[col sep=comma, x=x, y=deepbpd]{data/real_inference.csv};
                \addplot[name path=deepbpu,draw=none,fill=none] table[col sep=comma, x=x, y=deepbpu]{data/real_inference.csv};
                \addplot[cdeepBP,fill opacity=0.15] fill between[of=deepbpd and deepbpu];
                
                \addplot[name path=bpd,draw=none,fill=none] table[col sep=comma, x=x, y=bpd]{data/real_inference.csv};
                \addplot[name path=bpu,draw=none,fill=none] table[col sep=comma, x=x, y=bpu]{data/real_inference.csv};
                \addplot[cBP,fill opacity=0.15] fill between[of=bpd and bpu];
                
                \addplot[name path=deepmfd,draw=none,fill=none] table[col sep=comma, x=x, y=deepmfd]{data/real_inference.csv};
                \addplot[name path=deepmfu,draw=none,fill=none] table[col sep=comma, x=x, y=deepmfu]{data/real_inference.csv};
                \addplot[cdeepMF,fill opacity=0.15] fill between[of=deepmfu and deepmfd];
                
                \addplot[name path=mfd,draw=none,fill=none] table[col sep=comma, x=x, y=mfd]{data/real_inference.csv};
                \addplot[name path=mfu,draw=none,fill=none] table[col sep=comma, x=x, y=mfu]{data/real_inference.csv};
                \addplot[cMF,fill opacity=0.15] fill between[of=mfu and mfd];
                
                \addplot[name path=mvd,draw=none,fill=none] table[col sep=comma, x=x, y=mvd]{data/real_inference.csv};
                \addplot[name path=mvu,draw=none,fill=none] table[col sep=comma, x=x, y=mvu]{data/real_inference.csv};
                \addplot[cMV,fill opacity=0.15] fill between[of=mvu and mvd];
                
                \addplot[name path=mvd,draw=none,fill=none] table[col sep=comma, x=x, y=mvd]{data/real_inference.csv};
                \addplot[name path=mvu,draw=none,fill=none] table[col sep=comma, x=x, y=mvu]{data/real_inference.csv};
                \addplot[cMV,fill opacity=0.15] fill between[of=mvu and mvd];
                
                \addplot[name path=cld,draw=none,fill=none] table[col sep=comma, x=x, y=cld]{data/real_inference.csv};
                \addplot[name path=clu,draw=none,fill=none] table[col sep=comma, x=x, y=clu]{data/real_inference.csv};
                \addplot[cCL,fill opacity=0.15] fill between[of=clu and cld];
                
                \addplot[name path=bayesdgcd,draw=none,fill=none] table[col sep=comma, x=x, y=bayesdgcd]{data/real_inference.csv};
                \addplot[name path=bayesdgcu,draw=none,fill=none] table[col sep=comma, x=x, y=bayesdgcu]{data/real_inference.csv};
                \addplot[cBayesDGC,fill opacity=0.15] fill between[of=bayesdgcu and bayesdgcd];    
                
            \end{axis}
        \end{tikzpicture}
        \caption{Inference with $n$ spammers}
        \label{fig:real_world_extreme_inference}
    \end{subfigure} \hspace{-1.5mm} %
    \begin{subfigure}[t]{.24\linewidth}
        \centering
        \begin{tikzpicture}
            \begin{axis}[
                xlabel={$n$},
                ylabel={Test accuracy (\%)},
                width=1.2\linewidth,
                height=1.2\linewidth,
                xlabel style={yshift=0.15cm},
                ylabel style={yshift=-0.15cm},
                xmax=7.2,
                xmin=0.8
            ]
                \addplot[cdeepBP, very thick, mark=*, mark size=2pt, mark options={solid}] table[col sep=comma, x=x, y=deepbp]{data/real_learning.csv};
                \addplot[cdeepMF, very thick, mark=diamond*, mark size=2pt, mark options={solid}] table[col sep=comma, x=x, y=deepmf]{data/real_learning.csv};
                \addplot[cCL, very thick, mark=star, mark size=2pt, mark options={solid}] table[col sep=comma, x=x, y=cl]{data/real_learning.csv};
                \addplot[cBayesDGC, very thick, mark=x, mark size=2pt, mark options={solid}] table[col sep=comma, x=x, y=bayesdgc]{data/real_learning.csv};
                \addplot[cMV, very thick, mark=pentagon*, dashed, mark size=2pt, mark options={solid}] table[col sep=comma, x=x, y=mv]{data/real_learning.csv};
                
                \addplot[name path=deepbpd,draw=none,fill=none] table[col sep=comma, x=x, y=deepbpd]{data/real_learning.csv};
                \addplot[name path=deepbpu,draw=none,fill=none] table[col sep=comma, x=x, y=deepbpu]{data/real_learning.csv};
                \addplot[cdeepBP,fill opacity=0.15] fill between[of=deepbpd and deepbpu];
                
                \addplot[name path=deepmfd,draw=none,fill=none] table[col sep=comma, x=x, y=deepmfd]{data/real_learning.csv};
                \addplot[name path=deepmfu,draw=none,fill=none] table[col sep=comma, x=x, y=deepmfu]{data/real_learning.csv};
                \addplot[cdeepMF,fill opacity=0.15] fill between[of=deepmfu and deepmfd];
                
                \addplot[name path=cld,draw=none,fill=none] table[col sep=comma, x=x, y=cld]{data/real_learning.csv};
                \addplot[name path=clu,draw=none,fill=none] table[col sep=comma, x=x, y=clu]{data/real_learning.csv};
                \addplot[cCL,fill opacity=0.15] fill between[of=clu and cld];
                
                \addplot[name path=bayesdgcd,draw=none,fill=none] table[col sep=comma, x=x, y=bayesdgcd]{data/real_learning.csv};
                \addplot[name path=bayesdgcu,draw=none,fill=none] table[col sep=comma, x=x, y=bayesdgcu]{data/real_learning.csv};
                \addplot[cBayesDGC,fill opacity=0.15] fill between[of=bayesdgcu and bayesdgcd];
                
                \addplot[name path=mvd,draw=none,fill=none] table[col sep=comma, x=x, y=mvd]{data/real_learning.csv};
                \addplot[name path=mvu,draw=none,fill=none] table[col sep=comma, x=x, y=mvu]{data/real_learning.csv};
                \addplot[cMV,fill opacity=0.15] fill between[of=mvd and mvu];
            \end{axis}
        \end{tikzpicture}
        \caption{Learning with $n$ spammers}
        \label{fig:real_world_extreme_learning}
    \end{subfigure}
    \caption{{\em Face-age dataset experiments.} (a) The inference accuracy decreases as we blur the images more. (b) DeepBP outperforms the others in learning task when the images are relatively clear. (c, d) DeepBP is more robust than the other models against extreme-spammers added to the face-age dataset.}
    \label{fig:real_world_experiments}
\end{figure*}

\begin{table*}[th!]
    \centering
    \caption{{\em LabelMe experiments.} The performance is measured by training from scratch with 20 different seeds. DeepMF and deepBP outperform the other baseline models.}
    \begin{tabular}{l c c c c c}
        \toprule
        Models & MV & CL & BayesDGC & deepMF & deepBP \\
        \midrule
        Denoised accuracy (\%) & $77.22 \pm 0.10$ & $73.55 \pm 4.53$ & $77.91 \pm 0.43$ & $80.86 \pm 0.23$ & $\pmb{81.03 \pm 0.22}$ \\
        Test accuracy (\%) & $61.28 \pm 0.86$ & $63.38 \pm 3.81$ & $53.93 \pm 3.05$ & $\pmb{65.47 \pm 0.80}$ & $65.45 \pm 0.74$ \\
        \bottomrule
    \end{tabular}
    \label{tab:labelme}
\end{table*}

\paragraph{Overconfidence Issues.}
\label{sec:overconfidence}
We experimentally show that the BP-based algorithms are more robust than the MF-based algorithms on both inference and learning task under various scenarios. 
To understand the difference, we focus on the phenomenons which are known to appear in mean-field approximation: overconfidence and local minima~\citep{weiss2001comparing,murphy2013machine}.
In toy problems, \citet{weiss2001comparing} experimentally shows that the variational distributions of MF are likely to be overconfident and fall into local minima easily.

We investigate the problems in crowdsourcing systems. To understand overconfidence, we analyze the results used in \Figref{fig:extreme_spammer_mismatched_inference}. 
\Figref{fig:marginal} shows the distribution over marginal of true labels from the results. The marginals of MF skew to either zero or one, whereas those of BP distributed evenly around a half.
To understand how easily MF falls into a local minima, we run 100 experiments with different seeds and sort the inference accuracy in ascending order.
\Figref{fig:cvar} shows that there is a significant gap between MF and BP in the worst case, although their performance is not on average.
These two analyses show the overconfidence and local minima problems still exist in the crowdsourcing systems.
In addition, we investigate the distribution over the diagonal elements of the confusion matrices when MF trapped into a local minima. \Figref{fig:extreme_spammer_overconfidence} shows that MF estimates the extreme-spammer as an adversary, and therefore the labels of spammers misguide the inference steps in MF.
We conjecture that the overconfident makes MF-based algorithms sensitive but leave a thorough investigation for future work.




\subsection{Experiments on Real-world Datasets}
\label{sec:real_world}

For the real-world datasets, we use the face-age dataset from~\citet{han2015faceage} and LabelMe~\citep{rodrigues2018deep}.

\paragraph{Face-age.}
The face-age dataset is consisted of 1,002 human face images and their predicted ages answered by 165 workers from Amazon Mechanical Turk.
Since 20\% of the facial dataset are biased to infant images, we use 800 facial images over five-years-old.
We then transform the remaining dataset into a binary image classification that classifies whether a face is over 16-years-old to make a balanced classification problem.
The dataset is divided into 500 training and 300 test sets. 
On average, each worker labels 30 images in the original dataset. 
We generate a new assignment graph by randomly subsampling four tasks for each worker, i.e., $r=4$ to make the assignment structure sparse.
For the model prior, we use $\text{Dir}(1, 0.5)$ in \twofigref{fig:real_world_blur_inference}{fig:real_world_blur_learning}, and $\text{Dir}(2, 1.3)$ in \twofigref{fig:real_world_extreme_inference}{fig:real_world_extreme_learning}.
\Figref{fig:real_world_experiments} shows the performance of the algorithms with blurred images and extreme-spammers. The results show deepBP is robust against corrupted features and extreme-spammers in the real-world dataset as well.

\paragraph{LabelMe.}
LabelMe is a multi-class image classification task with eight classes consisted of 2,688 images.
We use 1,000 training images labeled by 59 workers from the Amazon Mechanical Turk.
The rest 1,688 images are used for testing.
We set the hyperparameter of prior $\alpha_{kk'}$ to 8 if $k = k'$ and 1 otherwise.
\autoref{tab:labelme} shows that both deepMF and deepBP outperform the existing methods.

\section{Conclusion}
In this work, we propose principled frameworks, named deepMF and deepBP, alternating variational inference and deep learning to utilize both the prior of worker behavior and the features of tasks.
We reveal that the previous deep crowdsourcing methods are special cases of deepMF with specific choices of worker prior. 
We propose deepBP inspired by the strong theoretical guarantee on BP for inference from crowds.
The experimental results show that the deepBP algorithm is more robust than the deepMF algorithm in canonical scenarios with non-informative features, multi-modal or mismatched worker prior, or the presence of extreme spammers.


\subsubsection*{Acknowledgments}
This work was partly supported by 
Institute of Information $\&$ communications Technology Planning $\&$ Evaluation (IITP) grant funded by the Korea government(MSIT) (No.2019-0-01906, Artificial Intelligence Graduate School Program(POSTECH)) 
and
Institute of Information $\&$ communications Technology Planning $\&$ Evaluation (IITP) grant funded by the Korea government(MSIT) (No.2021-0-02068, Artificial Intelligence Innovation Hub) 
and
the National Research Foundation of Korea(NRF) grant funded by the Korea government(MSIT) (No.2021R1C1C1011375)

\bibliography{references}

\begin{thebibliography}{}

\bibitem[Cao et~al., 2019]{cao2018maxmig}
Cao, P., Xu, Y., Kong, Y., and Wang, Y. (2019).
\newblock Max-{MIG}: an information theoretic approach for joint learning from
  crowds.
\newblock In {\em International Conference on Learning Representations}.

\bibitem[Chu and Wang, 2021]{chu2021generative}
Chu, Z. and Wang, H. (2021).
\newblock Improve learning from crowds via generative augmentation.
\newblock In {\em Proceedings of the 27th ACM SIGKDD Conference on Knowledge
  Discovery \& Data Mining}, KDD '21, page 167–175, New York, NY, USA.
  Association for Computing Machinery.

\bibitem[Dawid and Skene, 1979]{dawid1979maximum}
Dawid, A. and Skene, A. (1979).
\newblock Maximum likelihood estimation of observer error-rates using the em
  algorithm.
\newblock {\em Applied Statistics}, pages 20--28.

\bibitem[Gadiraju et~al., 2015]{gadiraju2015human}
Gadiraju, U., Demartini, G., Kawase, R., and Dietze, S. (2015).
\newblock Human beyond the machine: Challenges and opportunities of microtask
  crowdsourcing.
\newblock {\em IEEE Intelligent Systems}, 30(4):81--85.

\bibitem[Han et~al., 2015]{han2015faceage}
Han, H., Otto, C., Liu, X., and Jain, A.~K. (2015).
\newblock Demographic estimation from face images: Human vs. machine
  performance.
\newblock {\em IEEE Transactions on Pattern Analysis and Machine Intelligence},
  37(6):1148--1161.

\bibitem[Jagabathula et~al., 2017]{jagabathula2017identifying}
Jagabathula, S., Subramanian, L., and Venkataraman, A. (2017).
\newblock Identifying unreliable and adversarial workers in crowdsourced
  labeling tasks.
\newblock {\em The Journal of Machine Learning Research}, 18(1):3233--3299.

\bibitem[Kaggle, 2013]{dogsvscats}
Kaggle (2013).
\newblock Dogs vs. cats.
\newblock \url{https://www.kaggle.com/c/dogs-vs-cats/overview}.

\bibitem[Kuck et~al., 2020]{NEURIPS2020_07217414}
Kuck, J., Chakraborty, S., Tang, H., Luo, R., Song, J., Sabharwal, A., and
  Ermon, S. (2020).
\newblock Belief propagation neural networks.
\newblock In Larochelle, H., Ranzato, M., Hadsell, R., Balcan, M.~F., and Lin,
  H., editors, {\em Advances in Neural Information Processing Systems},
  volume~33, pages 667--678. Curran Associates, Inc.

\bibitem[Li et~al., 2021]{li2021crowdsourcing}
Li, S.-Y., Huang, S.-J., and Chen, S. (2021).
\newblock Crowdsourcing aggregation with deep bayesian learning.
\newblock {\em SCIENCE CHINA-INFORMATION SCIENCES}, 64(3).

\bibitem[Liu et~al., 2012]{liu2012variational}
Liu, Q., Peng, J., and Ihler, A. (2012).
\newblock Variational inference for crowdsourcing.
\newblock In {\em Proceedings of the 25th International Conference on Neural
  Information Processing Systems-Volume 1}, pages 692--700.

\bibitem[Murphy, 2013]{murphy2013machine}
Murphy, K.~P. (2013).
\newblock {\em Machine learning : a probabilistic perspective}.
\newblock MIT Press, Cambridge, Mass. [u.a.].

\bibitem[Nguyen et~al., 2017]{nguyen-etal-2017-aggregating}
Nguyen, A.~T., Wallace, B., Li, J.~J., Nenkova, A., and Lease, M. (2017).
\newblock Aggregating and predicting sequence labels from crowd annotations.
\newblock In {\em Proceedings of the 55th Annual Meeting of the Association for
  Computational Linguistics (Volume 1: Long Papers)}, pages 299--309,
  Vancouver, Canada. Association for Computational Linguistics.

\bibitem[Ok et~al., 2016]{ok2016optimality}
Ok, J., Oh, S., Shin, J., and Yi, Y. (2016).
\newblock Optimality of belief propagation for crowdsourced classification.
\newblock In {\em International Conference on Machine Learning}, pages
  535--544. PMLR.

\bibitem[Pearl, 1982]{pearl1982bp}
Pearl, J. (1982).
\newblock Reverend bayes on inference engines: A distributed hierarchical
  approach.
\newblock In {\em Proceedings of the Second AAAI Conference on Artificial
  Intelligence}, AAAI'82, page 133–136. AAAI Press.

\bibitem[Raykar et~al., 2010]{raykar2010learning}
Raykar, V.~C., Yu, S., Zhao, L.~H., Valadez, G.~H., Florin, C., Bogoni, L., and
  Moy, L. (2010).
\newblock Learning from crowds.
\newblock {\em Journal of Machine Learning Research}, 11(4).

\bibitem[Rodrigues and Pereira, 2018]{rodrigues2018deep}
Rodrigues, F. and Pereira, F. (2018).
\newblock Deep learning from crowds.
\newblock In {\em Proceedings of the AAAI Conference on Artificial
  Intelligence}, volume~32.

\bibitem[Rodrigues et~al., 2014]{rodrigues2014gaussian}
Rodrigues, F., Pereira, F., and Ribeiro, B. (2014).
\newblock Gaussian process classification and active learning with multiple
  annotators.
\newblock In {\em International conference on machine learning}, pages
  433--441. PMLR.

\bibitem[Russell et~al., 2008]{russell2008labelme}
Russell, B.~C., Torralba, A., Murphy, K.~P., and Freeman, W.~T. (2008).
\newblock Labelme: a database and web-based tool for image annotation.
\newblock {\em International journal of computer vision}, 77(1-3):157--173.

\bibitem[Satorras and Welling, 2021]{satorras2021neural}
Satorras, V.~G. and Welling, M. (2021).
\newblock Neural enhanced belief propagation on factor graphs.
\newblock In {\em International Conference on Artificial Intelligence and
  Statistics}, pages 685--693. PMLR.

\bibitem[Snow et~al., 2008]{snow2008cheap}
Snow, R., O’connor, B., Jurafsky, D., and Ng, A.~Y. (2008).
\newblock Cheap and fast--but is it good? evaluating non-expert annotations for
  natural language tasks.
\newblock In {\em Proceedings of the 2008 conference on empirical methods in
  natural language processing}, pages 254--263.

\bibitem[Tanno et~al., 2019]{tanno2019learning}
Tanno, R., Saeedi, A., Sankaranarayanan, S., Alexander, D.~C., and Silberman,
  N. (2019).
\newblock Learning from noisy labels by regularized estimation of annotator
  confusion.
\newblock In {\em Proceedings of the IEEE/CVF Conference on Computer Vision and
  Pattern Recognition}, pages 11244--11253.

\bibitem[Venanzi et~al., 2014]{venanzi2014community}
Venanzi, M., Guiver, J., Kazai, G., Kohli, P., and Shokouhi, M. (2014).
\newblock Community-based bayesian aggregation models for crowdsourcing.
\newblock In {\em Proceedings of the 23rd International Conference on World
  Wide Web}, WWW '14, page 155–164, New York, NY, USA. Association for
  Computing Machinery.

\bibitem[Weiss, 2001]{weiss2001comparing}
Weiss, Y. (2001).
\newblock Comparing the mean field method and belief propagation for
  approximate inference in mrfs.
\newblock {\em Advanced mean field methods: theory and practice}, pages
  229--240.

\bibitem[Welinder et~al., 2010]{welinder2010multidimensional}
Welinder, P., Branson, S., Perona, P., and Belongie, S. (2010).
\newblock The multidimensional wisdom of crowds.
\newblock {\em Advances in neural information processing systems},
  23:2424--2432.

\bibitem[Whitehill et~al., 2009]{whitehill2009whose}
Whitehill, J., Wu, T.-f., Bergsma, J., Movellan, J., and Ruvolo, P. (2009).
\newblock Whose vote should count more: Optimal integration of labels from
  labelers of unknown expertise.
\newblock {\em Advances in neural information processing systems},
  22:2035--2043.

\bibitem[Yedidia et~al., 2003]{yedidia2003understanding}
Yedidia, J.~S., Freeman, W.~T., Weiss, Y., et~al. (2003).
\newblock Understanding belief propagation and its generalizations.

\bibitem[Zhang et~al., 2020]{NEURIPS2020_61c66a2f}
Zhang, Z., Wu, F., and Lee, W.~S. (2020).
\newblock Factor graph neural networks.
\newblock In Larochelle, H., Ranzato, M., Hadsell, R., Balcan, M.~F., and Lin,
  H., editors, {\em Advances in Neural Information Processing Systems},
  volume~33, pages 8577--8587. Curran Associates, Inc.

\end{thebibliography}


\clearpage
\appendix

\thispagestyle{empty}

\onecolumn \makesupplementtitle

\section{Large but Sparse Dataset given Fixed Budget}
\label{appendix:fixed-budget}
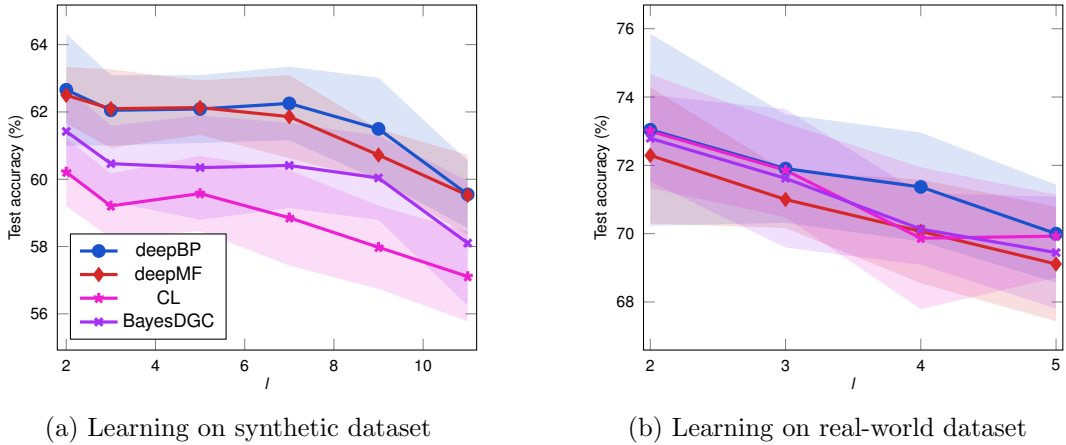
\begin{figure*}[!h]
    \captionsetup[subfigure]{font=normalsize}
    \centering
    \begin{subfigure}[t]{.45\linewidth}
        \centering
        \begin{tikzpicture}
            \begin{axis}[
                xlabel={$l$},
                ylabel={Test accuracy (\%)},
                xlabel style={yshift=0.15cm},
                ylabel style={yshift=-0.15cm},
                xmax=11.2,
                xmin=1.8,
                height=0.8\linewidth,
                legend style={nodes={scale=0.7, transform shape}},
                legend pos=south west,
            ]
                \addplot[cdeepBP, very thick, mark=*, mark size=2pt, mark options={solid}] table[col sep=comma, x=x, y=deepbp]{data/syn_budget.csv};
                \addplot[cdeepMF, very thick, mark=diamond*, mark size=2pt, mark options={solid}] table[col sep=comma, x=x, y=deepmf]{data/syn_budget.csv};
                \addplot[cCL, very thick, mark=star, mark size=2pt, mark options={solid}] table[col sep=comma, x=x, y=cl]{data/syn_budget.csv};
                \addplot[cBayesDGC, very thick, mark=x, mark size=2pt, mark options={solid}] table[col sep=comma, x=x, y=bayesdgc]{data/syn_budget.csv};
                
                \addplot[name path=deepbpd,draw=none,fill=none] table[col sep=comma, x=x, y=deepbpd]{data/syn_budget.csv};
                \addplot[name path=deepbpu,draw=none,fill=none] table[col sep=comma, x=x, y=deepbpu]{data/syn_budget.csv};
                \addplot[cdeepBP,fill opacity=0.15] fill between[of=deepbpd and deepbpu];
                
                \addplot[name path=deepmfd,draw=none,fill=none] table[col sep=comma, x=x, y=deepmfd]{data/syn_budget.csv};
                \addplot[name path=deepmfu,draw=none,fill=none] table[col sep=comma, x=x, y=deepmfu]{data/syn_budget.csv};
                \addplot[cdeepMF,fill opacity=0.15] fill between[of=deepmfu and deepmfd];
                
                \addplot[name path=cld,draw=none,fill=none] table[col sep=comma, x=x, y=cld]{data/syn_budget.csv};
                \addplot[name path=clu,draw=none,fill=none] table[col sep=comma, x=x, y=clu]{data/syn_budget.csv};
                \addplot[cCL,fill opacity=0.15] fill between[of=clu and cld];
                
                \addplot[name path=bayesdgcd,draw=none,fill=none] table[col sep=comma, x=x, y=bayesdgcd]{data/syn_budget.csv};
                \addplot[name path=bayesdgcu,draw=none,fill=none] table[col sep=comma, x=x, y=bayesdgcu]{data/syn_budget.csv};
                \addplot[cBayesDGC,fill opacity=0.15] fill between[of=bayesdgcu and bayesdgcd];
                
                \legend{deepBP,deepMF,CL,BayesDGC}
            \end{axis}
        \end{tikzpicture}
        \caption{Learning on synthetic dataset}
    \end{subfigure}
    \begin{subfigure}[t]{.45\linewidth}
        \centering
        \begin{tikzpicture}
            \begin{axis}[
                xlabel={$l$},
                ylabel={Test accuracy (\%)},
                xlabel style={yshift=0.15cm},
                ylabel style={yshift=-0.15cm},
                xmax=5.05,
                xmin=1.95,
                height=0.8\linewidth,
            ]
                \addplot[cdeepBP, very thick, mark=*, mark size=2pt, mark options={solid}] table[col sep=comma, x=x, y=deepbp]{data/real_budget.csv};
                \addplot[cdeepMF, very thick, mark=diamond*, mark size=2pt, mark options={solid}] table[col sep=comma, x=x, y=deepmf]{data/real_budget.csv};
                \addplot[cCL, very thick, mark=star, mark size=2pt, mark options={solid}] table[col sep=comma, x=x, y=cl]{data/real_budget.csv};
                \addplot[cBayesDGC, very thick, mark=x, mark size=2pt, mark options={solid}] table[col sep=comma, x=x, y=bayesdgc]{data/real_budget.csv};
                
                \addplot[name path=deepbpd,draw=none,fill=none] table[col sep=comma, x=x, y=deepbpd]{data/real_budget.csv};
                \addplot[name path=deepbpu,draw=none,fill=none] table[col sep=comma, x=x, y=deepbpu]{data/real_budget.csv};
                \addplot[cdeepBP,fill opacity=0.15] fill between[of=deepbpd and deepbpu];
                
                \addplot[name path=deepmfd,draw=none,fill=none] table[col sep=comma, x=x, y=deepmfd]{data/real_budget.csv};
                \addplot[name path=deepmfu,draw=none,fill=none] table[col sep=comma, x=x, y=deepmfu]{data/real_budget.csv};
                \addplot[cdeepMF,fill opacity=0.15] fill between[of=deepmfu and deepmfd];
                
                \addplot[name path=cld,draw=none,fill=none] table[col sep=comma, x=x, y=cld]{data/real_budget.csv};
                \addplot[name path=clu,draw=none,fill=none] table[col sep=comma, x=x, y=clu]{data/real_budget.csv};
                \addplot[cCL,fill opacity=0.15] fill between[of=clu and cld];
                
                \addplot[name path=bayesdgcd,draw=none,fill=none] table[col sep=comma, x=x, y=bayesdgcd]{data/real_budget.csv};
                \addplot[name path=bayesdgcu,draw=none,fill=none] table[col sep=comma, x=x, y=bayesdgcu]{data/real_budget.csv};
                \addplot[cBayesDGC,fill opacity=0.15] fill between[of=bayesdgcu and bayesdgcd];
            \end{axis}
        \end{tikzpicture}
        \caption{Learning on real-world dataset}
    \end{subfigure}
    \caption{{\em Fixed budget experiments.} Learning performance with varying $l$: 
    (a) 
    given $B = 2,000$, synthetic dataset generated with worker prior $\text{Dir}(2,1)$;
    and (b) given $B = 1,000$, the same real-world dataset used in Figure~\ref{fig:real_world_experiments}
    for which the algorithms use $\text{Dir}(2,1)$ as worker prior.
    }
    \label{fig:fixed_budget}
\end{figure*}

In our experiment, 
we mainly focus on the large-but-sparse dataset,
where the number of tasks $N$ is large but the number of workers per task is limited by a small constant $l$
due to budget constraint.
We often find such a dataset in practice, 
e.g., \citep{russell2008labelme, rodrigues2014gaussian},
although it is possible to use the same budget
for small-but-dense dataset with small $N$ and large $l$.
In this section, we formally study the trade-off between
large-but-sparse and small-but-dense datasets,
where, in turn, the crowdsoured dataset
is better to be large-but-sparse than
small-but-dense.
In \Figref{fig:fixed_budget}, 
we plot the performance of the set of learning algorithms
varying the choice of $N$ and $l$ given budget $B$
such that $N = \lfloor \frac{B}{l} \rceil$.
We observe that
regardless of which algorithm is employed,
the learning performance decreases as $l$ increases.
This suggests the large-but-sparse regime rather than
the small-but-dense one.

\vfill




\clearpage
\section{Details of Deep Mean-Field}
\label{appendix:deepMF}
We derive a detailed equation of deepMF alternating variational inference and learning descrbied in \Secref{sec:deepmf}.

DeepMF is dervied from an ELBO of the marginalization in \eqref{eq:two margin} by introducing a variational distribution $q(\rvz,\mTheta)$ in the followings:
\begin{equation}
\log p(\rvy \mid \rvx, \bm{\alpha}, \phi)
= \log \E_{q(\rvz, \mTheta)} \left[ \frac{p(\rvy, \rvz, \mTheta \mid \rvx, \bm{\alpha}, \phi)}{q(\rvz, \mTheta)} \right]
\geq \E_{q(\rvz, \mTheta)} \left[ \log \frac{p(\rvy, \rvz, \mTheta \mid \rvx, \bm{\alpha}, \phi)}{q(\rvz, \mTheta)} \right] \;.
\label{app:joint elbo}
\end{equation}
Note that the lower bound is maximized and achieves the equality 
if $q(\rvz,\mTheta) = p(\rvz, \mTheta \mid \rvx, \rvy, \bm\alpha, \phi)$.
We apply MF approximation,
assuming a conditional independence 
of $z_i$'s and $\mTheta^{(u)}$'s given $(\rvx, \rvy, \bm\alpha, \phi)$,
such that:
\begin{equation}
q(\rvz, \mTheta) = q(\rvz) q(\mTheta; \bm\beta) = 
\prod_{i\in[N]} q_i(z_i) \prod_{u\in[M]} 
q_u(\mTheta^{(u)}; \bm\beta^{(u)} ) \;,
\label{app:joint variational}
\end{equation}
where
we use a parametric estimation $q_u(\mTheta^{(u)}; \bm\beta^{(u)}) = \text{Dir} (\mTheta^{(u)}; \bm\beta^{(u)})$ for each worker $u$,
and 
approximating
$q_i(z_i) \approx p(z_i \mid \rvx, \rvy, \bm\alpha, \phi)$
and $q_u(\mTheta^{(u)}; \bm\beta^{(u)}) \approx p(\mTheta^{(u)} \mid \rvx, \rvy, \bm\alpha, \phi)$,
we let
$q(\rvz) = \prod_{i\in[N]} q_i(z_i)$ and $q(\mTheta;\bm\beta)  =\prod_{u\in[M]} q_u(\mTheta^{(u)};\bm\beta^{(u)})$.


Then, using \eqref{eqn:joint_learning} and \eqref{app:joint variational}, the ELBO in \eqref{app:joint elbo} is written as follows:
\begin{align}
\mathcal{L}_{\text{MF}}(q(\rvz) ; \phi, \mTheta, \bm\beta)
&:= \E_{q(\rvz) q(\mTheta;\bm\beta)} \left[ \log \frac{p(\rvy, \rvz, \mTheta \mid \rvx, \bm{\alpha}, \phi)}{q(\rvz) q(\mTheta;\bm\beta)} \right] \nonumber 
= \E_{q(\rvz) q(\mTheta;\bm\beta)} \left[ \log \frac{p(\rvz \mid \rvx, \phi)  p(\rvy \mid \rvz, \mTheta) p(\mTheta \mid \bm\alpha)}{q(\rvz) q(\mTheta;\bm\beta)} \right] \nonumber \\
&\;= \E_{q(\rvz) q(\mTheta; \bm\beta)} \left[ \log p(\rvy \mid \rvz, \mTheta) \right]
- \KL \left(q(\rvz) \parallel p(\rvz \mid \rvx, \phi)\right)
- \KL \left(q(\mTheta; \bm\beta) \parallel p(\mTheta \mid \bm{\alpha})\right) \;.
\label{app:deepmf}
\end{align}
In each iteration of deepMF,
we seek $q(\rvz)$, $\bm\beta$ and $\phi$, sequentially, 
to maximize $\mathcal{L}_{\text{MF}}(q(\rvz) ; \phi, \mTheta, \bm\beta)$
by fixing the others. To be specific, 
recalling $p(\rvz \mid \rvx, \phi) := \prod_{i \in [N]} f_\phi (z_i ; x_i)$,
given
the classifier $f_\phi$ and
the other variational distributions, 
the ELBO is maximized at $q_i(z_i)$ such that:
\begin{equation}
\log q_i (z_i)
= \sum_{u \in \sM_i} \E_{q_u(\mTheta^{(u)}; \bm\beta^{(u)})} \big[\log
\theta^{(u)}_{z_i, y_{i}^{(u)}}\big] 
+ \log f_\phi (z_i ; x_i) - 1 \;,
\label{app:inference mf}
\end{equation}
where we let $\sM_i$ be the set of workers labeling task $i$ 
and collection is denoted by $\text{MF} ({f}_{\phi}(\rvx), \rvy, \bm{\alpha}, \mTheta, \bm\beta )$.
The inference of deepMF
is based on $q_i(z_i)$ from $\text{MF}(\cdot)$.

Then, provided $q(\rvz)$ from $\text{MF}(\cdot)$, 
deepMF finds $\phi$ and $\bm\beta$ independently 
to maximize the ELBO in \eqref{app:deepmf}:
\begin{equation}
\hat{\phi}, \hat{\bm{\beta}} = \argmax_{\phi,\bm{\beta}} \mathcal{L}_{\text{MF}}(q(\rvz) ; \phi, \mTheta, \bm\beta) \;.
\end{equation}

First, the ELBO is maximized at $q_u(\mTheta^{(u)};\bm\beta^{(u)})$ such that:
\begin{equation}
\log q_u(\mTheta^{(u)};\bm\beta^{(u)})
= \sum_{i \in \sN_u} \E_{q(\rvz_i)} \left[\log p(y_{i}^{(u)} \mid \rvz_i, \mTheta^{(u)})\right] + \log p(\mTheta^{(u)} \mid \bm\alpha) - 1 \;,
\label{eq:concentration}
\end{equation}
where we use a worker prior $p(\mTheta^{(u)} \mid \bm\alpha)=\text{Dir}(\mTheta^{(u)};\bm\alpha)$ for each worker $u$. Each element of $\bm\beta^{(u)}$ is updated as:
\begin{equation}
\beta^{(u)}_{k_1k_2} = \alpha_{k_1k_2} + \sum_{\{i \mid i \in \sN_u, y^{(u)}_i = \, k_2\}}q_i(z_i=k_1) \;,
\end{equation}
for all $k_1, k_2 \in [K]$.

Second, the ELBO is maximized at $\phi$ such that:
\begin{equation}
\hat\phi = \argmax_{\phi} \mathcal{L}_{\text{MF}}(q(\rvz) ; \phi, \mTheta, \bm\beta)
= \argmin_{\phi} \KL(q(\rvz) \parallel p(\rvz \mid \rvx, \phi))\;,
\label{app:learning mf}
\end{equation}
which corresponds to training classifier $f_\phi$.

We note that 
deepMF
is easily implementable 
as
an iteration of the inference in \eqref{app:inference mf} and learning in \eqref{app:learning mf}
is just a sequence of maximizations for the same objective in \eqref{app:deepmf} but different variables.
Hence, it is a popular framework despite the coarse approximation in \eqref{app:joint variational}.

\vfill

\clearpage

\section{Correspondence of Deep Mean-Field to Existing Algorithms}
\label{appendix:MF-based algorithms}
\begin{table*}[h!]
    \setstretch{1.15}
    \centering
    \caption{{\em Choices of variational distributions and hyperparameter.} All the methods use fully-factorized variational distribution which follows MF approximation and Dirichlet distribution for worker prior $\bm\alpha$. To approximate the posterior of true label $q_i(z_i)$, CL and Trace use the output of a neural network $f_\phi(z_i;x_i)$. BayesDGC and deepMF estimate the Categorical distribution of $\bm\gamma$. For the posterior of confusion matrix $q_u(\mTheta^{(u)})$, BayesDGC and deepMF estimate the Dirichlet distribution of $\bm\beta$ to capture uncertainty. For the worker prior, CL set all $\bm\alpha$ to 1 which indicates no prior distribution on $\mTheta$. Trace posits an adversary prior that assumes more adversaries than hammers.
    }
    \begin{tabular}{l c c c}
        \toprule
        Model & $q_i(z_i)$ & $q_u(\mTheta^{(u)})$ & $\bm\alpha$ \\
        \midrule
        CL~\citep{rodrigues2018deep} & $f_{\phi}(z_i ; x_i)$ & $\delta(\mTheta=\mTheta^{(u)})$ & $\bm\alpha=\bm{1}$ \\
        Trace~\citep{tanno2019learning} & $f_{\phi}(z_i ; x_i)$ &$\delta(\mTheta=\mTheta^{(u)})$ & 
        $\alpha_{kk} < 1, \alpha_{kk'} = 1$ $(k'\neq k)$ \\
        BayesDGC~\citep{li2021crowdsourcing} & $\text{Cat}_i(z_i \mid \bm\gamma)$ & $\text{Dir}(\mTheta^{(u)} \mid \bm\beta^{(u)})$ & - \\
        deepMF & $\text{Cat}_i(z_i \mid \bm\gamma)$ & $\text{Dir}(\mTheta^{(u)} \mid \bm\beta^{(u)})$ & - \\
        \bottomrule
    \end{tabular}
    \label{table:special_cases}
\end{table*}

In \Secref{sec:connection}, we show that deepMF is a generalization of some previous studies based on choices of variational distributions and worker prior~\citep{rodrigues2018deep, tanno2019learning, li2021crowdsourcing}. In this section, we show that how these studies can be seen as a special case of deepMF. The choices of variational distributions and hyperparameters are summarized in \autoref{table:special_cases}.


First, the ELBO of deepMF in \eqref{eq:deepMF} is written as follows:
\begin{align}
\mathcal{L}_{\text{MF}}(q(\rvz) ; \phi, \mTheta, \bm\beta)
:= \E_{q(\rvz) q(\mTheta; \bm\beta)} [\log p(\rvy \mid \rvz, \mTheta)] 
- \KL(q(\rvz) \parallel p(\rvz | \rvx, \phi))
- \KL(q(\mTheta; \bm\beta) \parallel p(\mTheta | \bm{\alpha})) \;.
\label{eq:deepMF2}
\end{align}

Each \citet{rodrigues2018deep} and \citet{tanno2019learning} use a neural network parameterized by $\phi$ to model the posterior of true label $q(\rvz)$ and use the Dirac delta function as a variational distribution of confusion matrices $q(\mTheta)$ to simplify the training process. Then, the new ELBO is given by:
\begin{align}
\mathcal{L}_{\text{new}}(\rvz ; \phi, \mTheta)
:= \E_{p(\rvz \mid \rvx, \phi)} [\log p(\rvy \mid \rvz, \mTheta)]
+ \log p(\mTheta \mid \bm\alpha) \; .
\end{align}


The two algorithms choose different concentration parameters $\bm\alpha$ of the Dirichlet distribution. Assuming each confusion matrix $\mTheta^{(u)}$ is conditionally independent given $\bm\alpha$, the log-likelihood is written as:
\begin{equation}
\!\!\!\log p(\mTheta | \bm\alpha) \! = \!\!\sum_{u\in[M]} \!\!\log p(\mTheta^{(u)} | \bm\alpha)
= \!\!\sum_{u\in[M]} \sum_{k_1 \in [K]} \sum_{k_2 \in [K]} \!\!\log p(\theta^{(u)}_{k_1 k_2} | \alpha_{k_1 k_2})
\propto \!\sum_{u\in[M]} \sum_{k_1 \in [K]} \sum_{k_2 \in [K]} \!\!(\alpha_{k_1 k_2} - 1) \log \theta^{(u)}_{k_1 k_2} \!\;.
\end{equation}

\citet{rodrigues2018deep} set all $\bm{\alpha}$ to 1, which assumes that the workers are from a uniform distribution. Then, the ELBO of CL~\citep{rodrigues2018deep} is given by:
\begin{equation}
\mathcal{L}_{\text{CL}}(\rvz ; \phi, \mTheta)
:= \E_{p(\rvz \mid \rvx, \phi)} [\log p(\rvy \mid \rvz, \mTheta)] \;,
\end{equation}
where its negative represents the cross-entropy of the distribution $p(\rvy \mid \rvz, \mTheta)$ relative to the distribution $p(\rvz \mid \rvx, \phi)$.
\citet{tanno2019learning} set $\alpha_{k_1 k_2}$ less than 1 when $k_1 = k_2$ and 1 otherwise, which assumes that adversaries exist more than hammers, i.e., $\text{Dir}(0.8,1)$. Then, the ELBO of log version of Trace~\citep{tanno2019learning} is given by:
\begin{equation}
\mathcal{L}_{\text{Trace}}(\rvz ; \phi, \mTheta)
:= \E_{p(\rvz \mid \rvx, \phi)} [\log p(\rvy \mid \rvz, \mTheta)]
- \bm\lambda \text{ tr}(\log \mTheta) \;,
\label{eq:tanno}
\end{equation}
where $\bm\lambda$ is hyperparameter, i.e. $\bm\lambda \in \mathbb{R}^{K \times K} $ and $\lambda_{k_1 k_2} = 1 - \alpha_{k_1 k_2}$. $\text{tr}(\cdot)$ denotes trace and the difference from \citet{tanno2019learning} is the existence of log in the trace. The ELBO in \eqref{eq:tanno} simply add a regularization term of worker prior to CL \citep{rodrigues2018deep}.

However, the above uniform and adversary prior seem unreasonable in the real world. 
We find that the difference in prior is correlated with the initialization of the confusion matrix.
\citet{rodrigues2018deep} initialize the confusion matrix with identities, however, \citet{tanno2019learning} initialize the confusion matrix to totally experts, i.e., the diagonal component of the confusion matrix is greater than 1, for adversary prior, which can lower the value of the confusion matrix.

\vfill

\clearpage

\section{Sensitivity of Adversary Prior}
\label{appendix:tanno experiment}

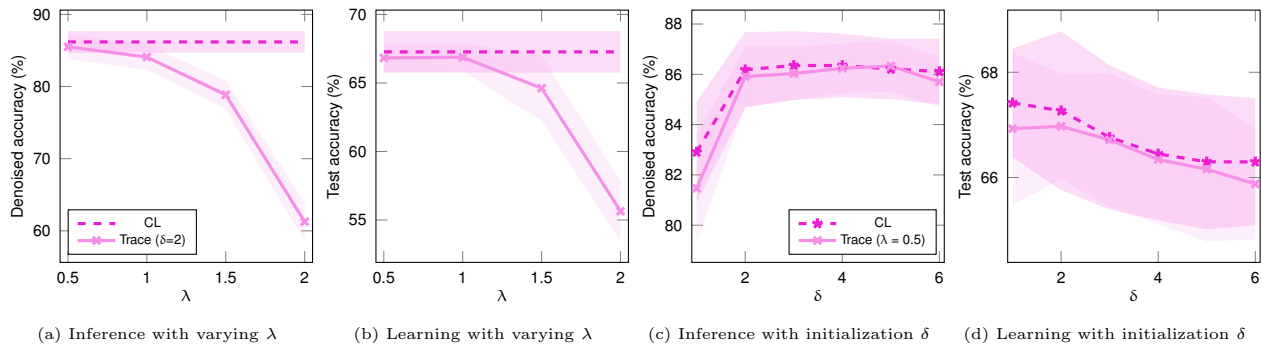
\begin{figure*}[!ht]
    \captionsetup[subfigure]{font=tiny,labelfont=tiny}
    \centering
    \begin{subfigure}[t]{.24\linewidth}
        \centering
        \begin{tikzpicture}
            \begin{axis}[
                legend style={nodes={scale=0.5}},
                xlabel={$\lambda$},
                ylabel={Denoised accuracy (\%)},
                width=1.2\linewidth,
                height=1.2\linewidth,
                xlabel style={yshift=0.15cm},
                ylabel style={yshift=-0.15cm},
                legend pos=south west,
                xmin=0.45,
                xmax=2.05
            ]
                \addplot[cCL, very thick, mark=none, dashed, mark size=2pt, mark options={solid}] table[col sep=comma, x=x, y=cl]{data/tanno_lambda_inference.csv};
                \addplot[cCL!50, very thick, mark=x, mark size=2pt, mark options={solid}] table[col sep=comma, x=x, y=tanno]{data/tanno_lambda_inference.csv};
                \legend{CL,Trace ($\delta$=2)}
                
                \addplot[name path=cld,draw=none,fill=none] table[col sep=comma, x=x, y=cld]{data/tanno_lambda_inference.csv};
                \addplot[name path=clu,draw=none,fill=none] table[col sep=comma, x=x, y=clu]{data/tanno_lambda_inference.csv};
                \addplot[cCL, fill opacity=0.15] fill between[of=clu and cld];
                
                \addplot[name path=tannod,draw=none,fill=none] table[col sep=comma, x=x, y=tannod]{data/tanno_lambda_inference.csv};
                \addplot[name path=tannou,draw=none,fill=none] table[col sep=comma, x=x, y=tannou]{data/tanno_lambda_inference.csv};
                \addplot[cCL!50, fill opacity=0.15] fill between[of=tannou and tannod];
            \end{axis}
        \end{tikzpicture}
        \caption{Inference with varying $\lambda$}
        \label{fig:tanno_lambda_inference}
    \end{subfigure} \hspace{-1.5mm} %
    \begin{subfigure}[t]{.24\linewidth}
        \centering
        \begin{tikzpicture}
            \begin{axis}[
                xlabel={$\lambda$},
                ylabel={Test accuracy (\%)},
                width=1.2\linewidth,
                height=1.2\linewidth,
                xlabel style={yshift=0.15cm},
                ylabel style={yshift=-0.15cm},
                xmin=0.45,
                xmax=2.05
            ]
                \addplot[cCL, very thick, mark=none, dashed, mark size=2pt, mark options={solid}] table[col sep=comma, x=x, y=cl]{data/tanno_lambda_learning.csv};
                \addplot[cCL!50, very thick, mark=x, mark size=2pt, mark options={solid}] table[col sep=comma, x=x, y=tanno]{data/tanno_lambda_learning.csv};
                
                \addplot[name path=cld,draw=none,fill=none] table[col sep=comma, x=x, y=cld]{data/tanno_lambda_learning.csv};
                \addplot[name path=clu,draw=none,fill=none] table[col sep=comma, x=x, y=clu]{data/tanno_lambda_learning.csv};
                \addplot[cCL, fill opacity=0.15] fill between[of=clu and cld];
                
                \addplot[name path=tannod,draw=none,fill=none] table[col sep=comma, x=x, y=tannod]{data/tanno_lambda_learning.csv};
                \addplot[name path=tannou,draw=none,fill=none] table[col sep=comma, x=x, y=tannou]{data/tanno_lambda_learning.csv};
                \addplot[cCL!50, fill opacity=0.15] fill between[of=tannou and tannod];
            \end{axis}
        \end{tikzpicture}
        \caption{Learning with varying $\lambda$}
        \label{fig:tanno_lambda_learning}
    \end{subfigure} \hspace{-1.5mm} %
    \begin{subfigure}[t]{.24\linewidth}
        \centering
        \begin{tikzpicture}
            \begin{axis}[
                legend style={nodes={scale=0.5}},
                legend pos=south east,
                xlabel={$\delta$},
                ylabel={Denoised accuracy (\%)},
                width=1.2\linewidth,
                height=1.2\linewidth,
                xlabel style={yshift=0.15cm},
                ylabel style={yshift=-0.15cm},
                xmin=0.9,
                xmax=6.1
            ]
                \addplot[cCL, very thick, mark=star, dashed, mark size=2pt, mark options={solid}] table[col sep=comma, x=x, y=cl]{data/tanno_init_inference.csv};
                \addplot[cCL!50, very thick, mark=x, mark size=2pt, mark options={solid}] table[col sep=comma, x=x, y=tanno]{data/tanno_init_inference.csv};
                \legend{CL,Trace ($\lambda=0.5$)}
                
                \addplot[name path=cld,draw=none,fill=none] table[col sep=comma, x=x, y=cld]{data/tanno_init_inference.csv};
                \addplot[name path=clu,draw=none,fill=none] table[col sep=comma, x=x, y=clu]{data/tanno_init_inference.csv};
                \addplot[cCL, fill opacity=0.15] fill between[of=clu and cld];
                
                \addplot[name path=tannod,draw=none,fill=none] table[col sep=comma, x=x, y=tannod]{data/tanno_init_inference.csv};
                \addplot[name path=tannou,draw=none,fill=none] table[col sep=comma, x=x, y=tannou]{data/tanno_init_inference.csv};
                \addplot[cCL!50, fill opacity=0.15] fill between[of=tannou and tannod];
            \end{axis}
        \end{tikzpicture}
        \caption{Inference with initialization $\delta$}
        \label{fig:tanno_init_inference}
    \end{subfigure} \hspace{-1.5mm} %
    \begin{subfigure}[t]{.24\linewidth}
        \centering
        \begin{tikzpicture}
            \begin{axis}[
                xlabel={$\delta$},
                ylabel={Test accuracy (\%)},
                width=1.2\linewidth,
                height=1.2\linewidth,
                xlabel style={yshift=0.15cm},
                ylabel style={yshift=-0.15cm},
                xmin=0.9,
                xmax=6.1
            ]
                \addplot[cCL, very thick, mark=star, dashed, mark size=2pt, mark options={solid}] table[col sep=comma, x=x, y=cl]{data/tanno_init_learning.csv};
                \addplot[cCL!50, very thick, mark=x, mark size=2pt, mark options={solid}] table[col sep=comma, x=x, y=tanno]{data/tanno_init_learning.csv};
                
                \addplot[name path=cld,draw=none,fill=none] table[col sep=comma, x=x, y=cld]{data/tanno_init_learning.csv};
                \addplot[name path=clu,draw=none,fill=none] table[col sep=comma, x=x, y=clu]{data/tanno_init_learning.csv};
                \addplot[cCL, fill opacity=0.15] fill between[of=clu and cld];
                
                \addplot[name path=tannod,draw=none,fill=none] table[col sep=comma, x=x, y=tannod]{data/tanno_init_learning.csv};
                \addplot[name path=tannou,draw=none,fill=none] table[col sep=comma, x=x, y=tannou]{data/tanno_init_learning.csv};
                \addplot[cCL!50, fill opacity=0.15] fill between[of=tannou and tannod];
            \end{axis}
        \end{tikzpicture}
        \caption{Learning with initialization $\delta$}
        \label{fig:tanno_init_learning}
    \end{subfigure}
    \caption{\textit{Sensitivity of Adversary Prior}. (a, b) The inference and learning performance of Trace decrease as $\lambda$ increases. We use confusion matrices initialized with $\delta=2$. (c, d) The inference and learning performance of both CL and Trace are sensitive to the initialization of confusion matrices. We set the hyperparameter $\lambda$ to 0.5. 
    }
    \label{fig:tanno_experiments}
\end{figure*}

Trace~\citep{tanno2019learning} proposes the adversary prior which assumes more adversaries than hammers. However, we empirically found the adversarial prior is sensitive to the choice of different hyperparameters. Specifically, we focuse on $\lambda$ and $\delta$, which indicates the degree of adversary and the initialization value of confusion matrices, respectively.
For the experiments, the confusion matrix is initialized as an diagonal matrix with the same diagonal entry $\delta$, e.g., $[[\delta, 0], [0, \delta]]$ for a binary classification.
We use a synthetic dataset generated with worker prior $\text{Dir}(3,1)$.
Different learning rates are set for classifier parameters and confusion matrix parameters: $1\mathrm{e}{-3}$ for training the classifier and $1\mathrm{e}{-2}$ for the confusion matrices, respectively. 
We conduct experiments from 7 different seeds.
As Trace constrains $\lambda$ to be positive, we vary $\lambda$ from 0.5 to 2. 
To initialize confusion matrices of workers to totally experts, we vary $\delta$ from 1 to 6.
\Figref{fig:tanno_experiments} shows that in both inference and learning task, the performance of Trace seems to be sensitive to $\lambda$ and $\delta$.
Even in the above situations, Trace performs worse than CL which assumes uniform prior.




\clearpage

\section{BP Optimization}
\label{appendix:bp_optimization}
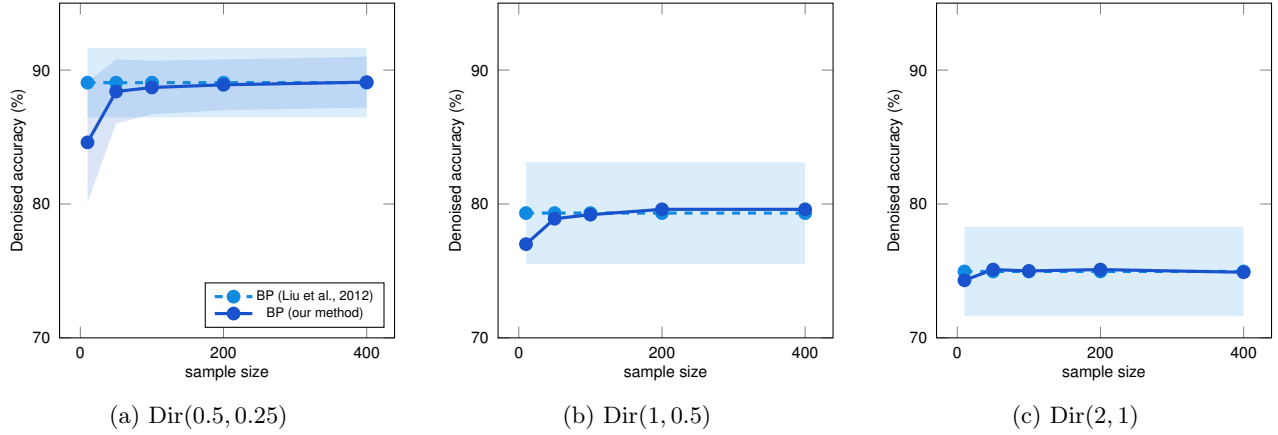
\begin{figure*}[h!]
    \centering
    \begin{subfigure}[t]{.32\linewidth}
        \centering
        \begin{tikzpicture}
            \begin{axis}[
                legend style={nodes={scale=0.50, transform shape}}, 
                table/col sep=comma,
                xlabel={sample size},
                ylabel={Denoised accuracy (\%)},
                legend pos=north east,
                width=1.1\linewidth,
                height=1.1\linewidth,
                legend pos=south east,
                xlabel style={yshift=0.15cm},
                ylabel style={yshift=-0.15cm},
                ymin=70,
                ymax=95
            ]
                \addplot[cBP, very thick, mark=*, dashed, mark size=2pt, mark options={solid}] table[x=samples, y=liu]{data/bp_performances_0_5.csv};
                \addplot[cdeepBP, very thick, mark=*, mark size=2pt, mark options={solid}] table[x=samples, y=bp]{data/bp_performances_0_5.csv};
                
                \addplot[name path=bpd,draw=none,fill=none] table[col sep=comma, x=samples, y=bpd]{data/bp_performances_0_5.csv};
                \addplot[name path=bpu,draw=none,fill=none] table[col sep=comma, x=samples, y=bpu]{data/bp_performances_0_5.csv};
                \addplot[cdeepBP,fill opacity=0.15] fill between[of=bpd and bpu];
                
                \addplot[name path=liud,draw=none,fill=none] table[col sep=comma, x=samples, y=liud]{data/bp_performances_0_5.csv};
                \addplot[name path=liuu,draw=none,fill=none] table[col sep=comma, x=samples, y=liuu]{data/bp_performances_0_5.csv};
                \addplot[cBP,fill opacity=0.15] fill between[of=liud and liuu];
                
                \legend{BP~\citep{liu2012variational}, BP (our method)};
            \end{axis}
        \end{tikzpicture}
        \caption{$\text{Dir}(0.5, 0.25)$}
    \end{subfigure}
    \hfill
    \begin{subfigure}[t]{.32\linewidth}
        \centering
        \begin{tikzpicture}
            \begin{axis}[
                legend style={nodes={scale=0.50, transform shape}}, 
                table/col sep=comma,
                xlabel={sample size},
                ylabel={Denoised accuracy (\%)},
                width=1.1\linewidth,
                height=1.1\linewidth,
                xlabel style={yshift=0.15cm},
                ylabel style={yshift=-0.15cm},
                ymin=70,
                ymax=95
            ]
                \addplot[cBP, very thick, mark=*, dashed, mark size=2pt, mark options={solid}] table[x=samples, y=liu]{data/bp_performances_1.csv};
                \addplot[cdeepBP, very thick, mark=*, mark size=2pt, mark options={solid}] table[x=samples, y=bp]{data/bp_performances_1.csv};
                
                \addplot[name path=liud,draw=none,fill=none] table[col sep=comma, x=samples, y=liud]{data/bp_performances_1.csv};
                \addplot[name path=liuu,draw=none,fill=none] table[col sep=comma, x=samples, y=liuu]{data/bp_performances_1.csv};
                \addplot[cBP,fill opacity=0.15] fill between[of=liud and liuu];
            \end{axis}
        \end{tikzpicture}
        \caption{$\text{Dir}(1, 0.5)$}
    \end{subfigure}
    \hfill
    \begin{subfigure}[t]{.32\linewidth}
        \centering
        \begin{tikzpicture}
            \begin{axis}[
                legend style={nodes={scale=0.50, transform shape}}, 
                table/col sep=comma,
                xlabel={sample size},
                ylabel={Denoised accuracy (\%)},
                width=1.1\linewidth,
                height=1.1\linewidth,
                xlabel style={yshift=0.15cm},
                ylabel style={yshift=-0.15cm},
                ymin=70,
                ymax=95
            ]
                \addplot[cBP, very thick, mark=*, dashed, mark size=2pt, mark options={solid}] table[x=samples, y=liu]{data/bp_performances_2.csv};
                \addplot[cdeepBP, very thick, mark=*, mark size=2pt, mark options={solid}] table[x=samples, y=bp]{data/bp_performances_2.csv};
                
                \addplot[name path=liud,draw=none,fill=none] table[col sep=comma, x=samples, y=liud]{data/bp_performances_2.csv};
                \addplot[name path=liuu,draw=none,fill=none] table[col sep=comma, x=samples, y=liuu]{data/bp_performances_2.csv};
                \addplot[cBP,fill opacity=0.15] fill between[of=liud and liuu];
            \end{axis}
        \end{tikzpicture}
        \caption{$\text{Dir}(2, 1)$}
    \end{subfigure}
    \caption{\textit{Inference performance of our BP and \citet{liu2012variational}'s BP}. We compare the inference performance of our BP algorithm and \citet{liu2012variational}'s BP algorithm in three different worker priors. The inference performance is averaged from 100 different seeds. (a, b, c) In all the settings, as the sample size increases, our BP's inference performance is improved and shows similar performance to \citet{liu2012variational}. We use sample size 400.}
    \label{fig:bp_performances}
\end{figure*}

\subsection{Factor-to-Variable Message}
We remark that the update of $m^{t+1}_{g_u \rightarrow i}$ in \eqref{eq:worker_to_task}
seems intractable as it requires exponentially many summations in the number of tasks labeled by worker, i.e., $O(2^{|\sN_u|})$.
To bypass the intractable computation,
we use a Monte-Carlo method using $S$ samples,
of which computational cost is bounded by $O(|\sN_u| \cdot K \cdot S)$.

\paragraph{Expansion.}
The update of factor-to-variable message in \eqref{eq:worker_to_task} can be written as:
\begin{align}
m_{g_u \rightarrow i}(z_i) &\propto \sum_{\rvz_{\sN_u \backslash \{i\}}} g_u (\rvz_{\sN_u}) \prod_{j \in \sN_u \backslash i} m_{j \rightarrow g_u}(z_j) \\
&= \sum_{\rvz_{\sN_u \backslash \{i\}}} \left(\int_{\mTheta^{(u)}} p(\mTheta^{(u)} \mid \bm{\alpha}) \prod_{(k_1, k_2) \in [K] \times [K]} ( \theta^{(u)}_{k_1, k_2} )^{\gamma_{k_1k_2}^{(u)}} \, d\mTheta^{(u)}\right) \prod_{j \in \sN_u \backslash i} m_{j \rightarrow g_u}(z_j) \\
&= \int_{\mTheta^{(u)}} p(\mTheta^{(u)} \mid \bm{\alpha}) \cdot \sum_{\rvz_{\sN_u \backslash \{i\}}} \left( \prod_{(k_1, k_2) \in [K] \times [K]} (\theta^{(u)}_{k_1, k_2})^{\gamma_{k_1k_2}^{(u)}} \prod_{j \in \sN_u \backslash i} m_{j \rightarrow g_u}(z_j) \right) \, d\mTheta^{(u)} \\
&= \int_{\mTheta^{(u)}} p(\mTheta^{(u)} \mid \bm\alpha) \cdot \theta^{(u)}_{z_i, y^{(u)}_i} \cdot \sum_{\rvz_{\sN_u \backslash i}} \prod_{j \in \sN_u \backslash i} \theta^{(u)}_{z_j, y^{(u)}_j} \cdot m_{j \rightarrow g_u}(z_j) \, d\mTheta^{(u)} \\
&=\int_{\mTheta^{(u)}} p(\mTheta^{(u)} \mid \bm{\alpha}) \cdot \theta^{(u)}_{z_i, y_i^{(u)}} \cdot \prod_{j \in \sN_u \backslash \{i\}} \sum_{z \in [K]} \theta^{(u)}_{z, y_j^{(u)}} \cdot m_{j \rightarrow g_u}(z) \, d\mTheta^{(u)} \\
&=\int_{\mTheta^{(u)}} p(\mTheta^{(u)} \mid \bm{\alpha}) \cdot \theta^{(u)}_{z_i, y_i^{(u)}} \cdot \prod_{j \in \sN_u \backslash i} \langle \theta^{(u)}_{\cdot, y_j^{(u)}}, \rvm_{j \rightarrow g_u} \rangle \, d\mTheta^{(u)} \\
&= \E_{p(\mTheta^{(u)} \mid \bm\alpha)}\left[ \theta^{(u)}_{z_i, y_i^{(u)}} \cdot \prod_{j \in \sN_u \backslash i} \langle \theta^{(u)}_{\cdot, y_j^{(u)}}, \rvm_{j \rightarrow g_u} \rangle \right] \;, \label{eq:our_bp}
\end{align}
where $\gamma^{(u)}_{k_1 k_2}$ is the number tasks that the true label is $k_1$ and the worker $u$'s label is $k_2$, and $\rvm_{j \rightarrow g_u}$ is a vector of the messages from task $j$ to worker $u$ where $\rvm_{j \rightarrow g_u} \in \R^K$. 
By approximating $p(\mTheta^{(u)} \mid \bm\alpha)$ with $S$ Monte-Carlo samples, we can compute the expectation \eqref{eq:our_bp} in $O(|\sN_u| \cdot K \cdot S)$.

\subsection{Effect of the Sample Size}

As our BP algorithm in \eqref{eq:our_bp} uses Monte-Carlo sampling, its performance is inconsistent when using continuous worker prior, for example, Beta and Dirichlet.
To provide a performance boundary, we test our BP algorithm in synthetic datasets, where the true prior is continuous, with varying sample size from 10 to 400 and compare it with \citet{liu2012variational}'s BP algorithm when the true prior is given to the algorithms.
Each of the synthetic dataset is generated with worker prior $\text{Dir}(2, 1)$, $\text{Dir}(1, 0.5)$, and $\text{Dir}(0.5, 0.25)$. 
\Figref{fig:bp_performances} shows that in all the settings, the inference performance of our BP algorithm is improved as the sample size increases.
Furthermore, with many samples, it shows comparable performance to \citet{liu2012variational}.

\clearpage

\section{Experimental Details}
\begin{table}[h]
    \centering
    \caption{\textit{Learning rates of the experiments}. CL denotes \citet{rodrigues2018deep} and BayesDGC denotes \citet{li2021crowdsourcing}.}
    \begin{tabular}{l c c c c c}
        \toprule
        \textbf{Models} & \textbf{MV} & \textbf{CL} & \textbf{BayesDGC} & \textbf{deepMF} & \textbf{deepBP} \\
        \midrule
        \textbf{\Figref{fig:feature_clip}} & - & - & - & $1\mathrm{e}{-4}$ & $1\mathrm{e}{-4}$ \\
        \textbf{\Figref{fig:true_prior}} & $1\mathrm{e}{-4}$ & $1\mathrm{e}{-4}$ & $1\mathrm{e}{-4}$ & $1\mathrm{e}{-4}$ & $1\mathrm{e}{-4}$ \\
        \textbf{Figures \ref{fig:extreme_spammer_mismatched_inference}, \ref{fig:extreme_spammer_mismatched_learning}} & $1\mathrm{e}{-4}$ & $1\mathrm{e}{-4}$ & $1\mathrm{e}{-5}$ & $1\mathrm{e}{-4}$ & $1\mathrm{e}{-4}$ \\
        \textbf{Figures \ref{fig:extreme_spammer_frequent_inference}, \ref{fig:extreme_spammer_frequent_learning}} & $1\mathrm{e}{-4}$ & $1\mathrm{e}{-4}$ & $1\mathrm{e}{-4}$ & $1\mathrm{e}{-4}$ & $1\mathrm{e}{-4}$ \\
        \textbf{Figures \ref{fig:real_world_blur_inference}, \ref{fig:real_world_blur_learning}} & $1\mathrm{e}{-3}$ & $1\mathrm{e}{-3}$ & $1\mathrm{e}{-3}$ & $1\mathrm{e}{-3}$ & $1\mathrm{e}{-3}$ \\
        \textbf{Figures \ref{fig:real_world_extreme_inference}, \ref{fig:real_world_extreme_learning}} & $5\mathrm{e}{-5}$ & $5\mathrm{e}{-5}$ & $5\mathrm{e}{-5}$ & $5\mathrm{e}{-5}$ & $5\mathrm{e}{-5}$ \\
        \textbf{Table \ref{tab:labelme}} &$1\mathrm{e}{-4}$&$1\mathrm{e}{-3}$&$1\mathrm{e}{-4}$&$1\mathrm{e}{-4}$&$1\mathrm{e}{-4}$ \\
        \textbf{Figures \ref{fig:fixed_budget}} & - & $1\mathrm{e}{-4}$ & $1\mathrm{e}{-4}$ & $1\mathrm{e}{-4}$ & $1\mathrm{e}{-4}$ \\
        \bottomrule
    \end{tabular}
    \label{table:learning_rates}
\end{table}

\label{appendix:exp-details}
\subsection{Model Architecture}
Due to the complexity between the input features and the true labels, the classifier in \eqref{eqn:task_model} is often modeled with neural networks. 
We follow the standard architecture of neural network in crowdsourcing literature for fair comparison, e.g., \citep{rodrigues2018deep} and \citep{cao2018maxmig}. For the classifier, we use a four-layer convolutional network with kernel size 3 and stride 1 without padding, and the number of output channels for each convolutional layer is 32, 32, 64, and 64.
Each of the convolutional layer's output passes through a max-pooling layer with kernel size 2 and stride 2.
Batch normalization is used at the output of the first convolutional layer.
By passing the last convolutional layer's output to two fully-connected  layers with hidden size 128 and ReLU activation, we finally retrieve the prediction of the input image.

\subsection{Hyperparameters}
We train the classifier with full-batch to maintain the whole structure of the given assignment graph in all the experiments.
For approximating the factor-to-variable messages in \eqref{eq:our_bp}, we use sample size 400.
We clip the classifier's output with value 0.9 for all the experiments, except \Figref{fig:feature_clip}.
\autoref{table:learning_rates} shows the learning rates for each experiment.
For the computing infrastructures, we use RTX 3090 and RTX A5000.


\clearpage

\section{Examples of Blurred Images}
\label{appendix:blur-images}
\begin{figure*}[h!]
    \captionsetup[subfigure]{font=normalsize}
    \centering
    \begin{subfigure}[h!]{.32\linewidth}
        \centering
        \includegraphics[scale=0.75]{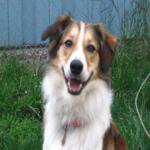}
        \caption{0 blur radius}
    \end{subfigure}
    \hspace{-1mm}
    \begin{subfigure}[h!]{.32\linewidth}
        \centering
        \includegraphics[scale=0.75]{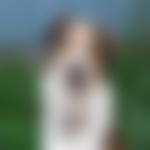}
        \caption{8 blur radius}
    \end{subfigure}
    \hspace{-1mm}
    \begin{subfigure}[h!]{.32\linewidth}
        \centering
        \includegraphics[scale=0.75]{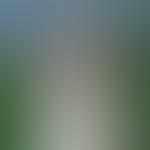}
        \caption{32 blur radius}
        \label{fig:blur 32 dog}
    \end{subfigure}
    \begin{subfigure}[h!]{.32\linewidth}
        \centering
        \includegraphics[scale=0.75]{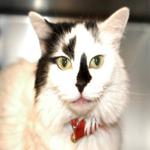}
        \caption{0 blur radius}
    \end{subfigure}
    \hspace{-1mm}
    \begin{subfigure}[h!]{.32\linewidth}
        \centering
        \includegraphics[scale=0.75]{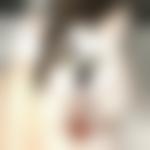}
        \caption{8 blur radius}
    \end{subfigure}
    \hspace{-1mm}
    \begin{subfigure}[h!]{.32\linewidth}
        \centering
        \includegraphics[scale=0.75]{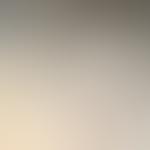}
        \caption{32 blur radius}
        \label{fig:blur 32 cat}
    \end{subfigure}
    \caption{{\em Blurred images.} (a, b, c) Dog images with blur radius from 0 to 32. (d, e, f) Cat images with blur radius from 0 to 32. When blur radius reaches 32, it is tricky to distinguish between dogs and cats with the naked eyes.}
    \label{fig:blur image}
\end{figure*}


In \Secref{sec:robustness to feature}, we show the performance of deepMF and deepBP when the task features are corruped by the Gaussian blur.
To exemplify the effect of Gaussian blur, we demonstrate the changes in images with varying radius of Gaussian blur in \Figref{fig:blur image}.
When the radius of Gaussian blur reaches 32, it is almost impossible to identify the original class of the image by naked eyes as shown in \Figref{fig:blur 32 dog} and \Figref{fig:blur 32 cat}.

\newpage
\section{Existence of Extreme-spammers}
\label{appendix:dataset-details}
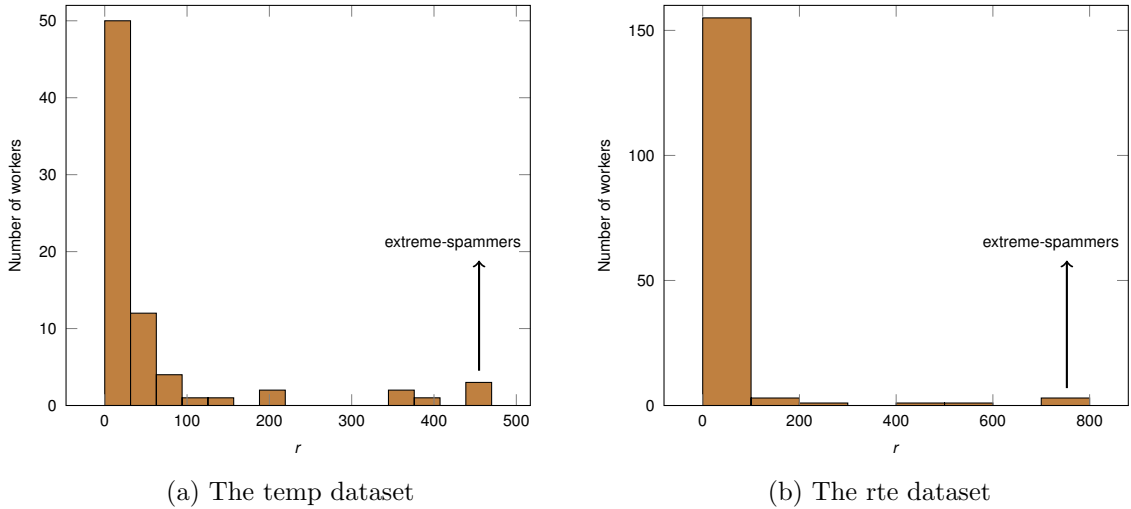
\begin{figure*}[h!]
    \captionsetup[subfigure]{font=normalsize}
    \begin{center}
    \begin{subfigure}{.45\linewidth}
        \begin{tikzpicture}
            \begin{axis}[
                area style,
                scale only axis,
                ymin=0,
                legend pos=north west,
                width=0.8\linewidth,
                ymax=52,
                xlabel={$r$},
                ylabel={Number of workers}
            ]
                \addplot+[hist={bins=15, data max=470, data min=0}, fill=brown, draw=black, thin] table[col sep=comma, y=temp]{data/dataset_hist.csv};
                \draw [->, thick] (0.645\linewidth, 0.06\linewidth)--++(0, 0.19\linewidth);
                \draw (0.60\linewidth, 0.28\linewidth) node {\tiny\textsf{extreme-spammers}};
            \end{axis}
        \end{tikzpicture}
        \caption{The temp dataset}
        \label{fig:temp}
    \end{subfigure}
    \begin{subfigure}{.45\linewidth}
        \begin{tikzpicture}
            \begin{axis}[
                area style,
                scale only axis, 
                ymin=0,
                legend pos=north west,
                ymax=160,
                ylabel={Number of workers},
                xlabel={$r$},
                width=0.8\linewidth
            ]
                \addplot+[hist={bins=8, data max=800, data min=0}, fill=brown, draw=black, thin] table[col sep=comma, y=rte]{data/dataset_hist.csv};
                \draw [->, thick] (0.6275\linewidth, 0.03\linewidth)--++(0, 0.22\linewidth);
                \draw (0.60\linewidth, 0.28\linewidth) node {\tiny\textsf{extreme-spammers}};
            \end{axis}
        \end{tikzpicture}
        \caption{The rte dataset}
        \label{fig:rte}
    \end{subfigure}
    \end{center}
    \caption{{\em Existence of extreme-spammers.} (a) The temp dataset~\citep{snow2008cheap} consists of 462 tasks and 76 workers. Three extreme-spammers solve 442, 452, and 462 tasks with correct answer rate 0.50, 0.44, and 0.44, respectively. (b) The rte dataset~\citep{snow2008cheap} consists of 800 tasks and 164 workers. Four extreme-spammers solve 540, 700, 760, and 800 tasks with correct answer rate 0.50, 0.58, 0.49, and 0.50, respectively. We observe that extreme-spammers exists in the real-world datasets.}
    \label{fig:dataset}
\end{figure*}

BP-based algorithms, such as BP and deepBP, are robust to extreme-spammer who labels uniformly across all tasks than MF-based algorithms as show in  \Secref{sec:extreme-spammer}.
To show that these extreme-spammers not only exist in the imaginary scenarios, we investigate the existence of extreme-spammers in real-world datasets.
\Figref{fig:dataset} describes histograms of $r$, the number of tasks per worker, from the temp dataset~\citep{snow2008cheap} and the rte dataset~\citep{snow2008cheap}. In this figure, the workers who attempt the most of the tasks are identified as an extreme-spammers. Specifically, we can identify three and four extreme-spammers in the temp and the rte dataset, respectively.
As we have shown in the main experiments, a few number of extreme-spammer can mis-guide the entire inference and learning algorithms. Therefore, it is important to have robust algorithms working with these extreme-spammers.

\clearpage

\end{document}